\documentclass[]{article}

\usepackage{jmlr2e}

\RequirePackage[OT1]{fontenc}
\RequirePackage{amsmath}
\usepackage{amsfonts} 
\usepackage{algorithm,algpseudocode}
\usepackage{ulem}
\usepackage{xcolor}
\RequirePackage{subcaption}

\newtheorem{assumption}{Assumption}

\input{Definitions}

\newcommand{\bsx}{{\boldsymbol{x}}}
\newcommand{\bsu}{{\boldsymbol{u}}}

\newcommand{\bsv}{{\boldsymbol{v}}}
\newcommand{\bsy}{{\boldsymbol{y}}}
\newcommand{\bsz}{{\boldsymbol{z}}}
\newcommand{\bsf}{{\boldsymbol{f}}}
\newcommand{\btheta}{{\boldsymbol \theta}}
\newcommand{\cx}{{\cal X}}

\newcommand{\real}{{\mathbb R}}
\newcommand{\bignorm}[1]{\left\lVert#1\right\rVert}

\begin{document}

\title{Adaptive Rate of Convergence of Thompson Sampling for Gaussian Process Optimization}

\author{\name Kinjal Basu \email kbasu@linkedin.com \\
       \addr 700 E Middlefield Road,\\
       Mountain View, CA 94043, USA
       \AND
       \name Souvik Ghosh \email sghosh@linkedin.com \\
       \addr 
       700 E Middlefield Road,\\
       Mountain View, CA 94043, USA}

\editor{}

\maketitle

\begin{abstract}
We consider the problem of global optimization of a  function over a continuous domain. In our setup, we can evaluate the function sequentially at points of our choice and the evaluations are noisy.    We frame it as a continuum-armed bandit problem with a Gaussian Process prior on the function. In this regime, most algorithms have been developed to minimize some form of regret. In this paper, we study the convergence of the sequential point $\bsx^t$ to the global optimizer $\bsx^*$ for the Thompson Sampling approach. Under some assumptions and regularity conditions, we prove concentration bounds for $\bsx^t$  where the probability that $\bsx^t$  is bounded away from $\bsx^*$  decays exponentially fast in $t$. Moreover, the result allows us to derive adaptive convergence rates depending on the function structure.   
\end{abstract}


\section{Introduction}
\label{sec:intro}

Let $f : \cx \rightarrow \real$ be an unknown function defined on a compact set $\cx \subset \real^d$. We are interested in solving the global maximization problem and obtaining the global maximizer 
\[\bsx^* = \argmax_{\bsx \in \cx} f(\bsx).
 \] 
We assume that the space $\cx$ is continuous and that $\bsx^*$ is unique, i.e. the function $f$ has a unique global maximizer.  

Such optimization problems are common in scientific and engineering fields. Examples include learning continuous valuation models \citep{eric2008active}, automatic gait optimization for both quadrupedal and bipedal robots \citep{lizotte2007automatic},  choosing the optimal derivative of a molecule that best treats a disease \citep{negoescu2011knowledge}, tuning Hamiltonian based Monte Carlo Samplers \citep{wang2013adaptive}, etc. A good survey of the problem in practical machine learning applications is presented in \cite{snoek2012practical}.  
Our motivation for studying this problem stems from an application of  recommender systems, where the goal is to rank multiple types of items like articles, videos, ads and jobs on a webpage to optimize a diverse range of business metrics like user engagement,  revenue from advertisements and job applications from job seekers. In our example, the function $f(\bsx)$ is a utility function composed of various business metrics and $\bsx$ are parameters or knobs that control the relative frequency of different types of items we show on the webpage. See \cite{oms2018} for more details.

Global optimization of such functions  is close to impossible without any further assumption on $f$. It is common to assume a Gaussian Process (GP) prior on the function $f$. These assumptions help formulate algorithms such as GP-UCB and its variants for explore-exploit. Many such variants have been well studied \citep{auer2002finite, garivier2011kl, hernandez2014predictive, kaufmann2012bayesian, lai1985asymptotically, maillard2011finite}. Some theoretical properties are also known for such algorithms \citep{srinivas2009gaussian, srinivas2012information}. 

The main idea is to optimize an  \emph{acquisition function} to determine the next point where we evaluate the function. Most analyses of such algorithms give an upper bound to the average \emph{cumulative regret},
\begin{equation}
\label{regret}
\frac{R_T}{T} = \frac{1}{T} \sum_{t=1}^T \left(f(\bsx^*) - f(\bsx_t)\right).
\end{equation}
\cite{Kandasamy:2016:GPB:3157096.3157208} derive regret guarantees for GP optimization algorithms (even in the more general multi-fidelity
setting). For other streams of work around GP-UCB and its variants see \cite{Valko:2013:FAK:3023638.3023705, Chowdhury:2017:KMB:3305381.3305469}.

In this paper, we focus on an approach known as Thompson Sampling (TS). Although this is an old idea dating back to \cite{thompson1933likelihood}, there has been considerable attention in the recent past \citep{bijl2016sequential, granmo2010solving, may2011simulation}. Studies have shown good empirical evidence of efficiency of TS \citep{chapelle2011empirical} and more recently, theoretical proofs have been obtained for the multi-arm bandit setting and some generalizations. \citet{agrawal2012analysis} showed for the first time that TS achieves logarithmic expected regret for the stochastic multi-armed bandit problem.   The same authors \citep{agrawal2013further} provided a near-optimal bound of $O(\sqrt{NT \log T})$ for expected regret of TS for the $N$-armed bandit problem. \citet{agrawal2013thompson} gave further results on contextual multi-armed bandits with linear payoffs. Analysis for the infinite armed bandit on a continuous space was missing, until \citet{russo2014learning} gave an overview of how to bound the regret by drawing an analogy between TS and Upper Confidence Bound (UCB) algorithms. 

Throughout this paper, we assume that we can evaluate the function sequentially and that the function evaluations are noisy. At every attempt $t$, we choose $\bsx^t \in \cx$ and observe $y^t = f(\bsx^t) + \epsilon^t$, where $\epsilon^t$ are independent errors in each observation with $\epsilon^t \sim N(0, \sigma^2)$ and $\sigma^2$ is unknown (but fixed). In many applications, $y^t$ is the reward reaped with attempt $t$ and the goal is to maximize reward over time. This naturally leads to  \emph{explore-exploit} type of algorithms and regret analysis for such algorithms. In many other applications though, there are no obvious notion of regret and finding the optima is more important. In such examples a guarantee on the convergence of the sequential point $\bsx^t$ to the true global optima $\bsx^*$ are important. This convergence phenomenon is our focus in this paper. 

Our motivating example is the problem of ranking news items on a social network feed. \cite{oms2018} present such an application in good details. A social network feed typically comprises of different types of items like news articles, videos, ads, jobs etc. A common engineering framework to support such a feed involve multiple individual systems that are responsible for generating a specific type of content and another system (blender) that blends the different items together. The blender typically tries to balance various business metrics like user engagement, revenue, job applications etc. A common development model would involve separate teams who own the individual systems and develop independently. This is a challenge for the blender system as it has to frequently adapt to the changes made by the individual systems.  Some changes in an individual system might adversely affect others- for example, if the video recommendation algorithm improves then videos might start to show up more in higher ranks and hence displacing ads and reducing revenue from ads. In such situations the blender can adapt by reducing the relative frequency of videos to get revenue back to the previous levels. We can formulate the function of the blender as an optimization problem where for any change from an individual system, the blender tunes certain parameters (eg. vector of frequencies of different types of items) so that it gets the maximum benefit possible for the individual system while not affecting other metrics. A natural question is how quickly can the blender find the new balance. 

Another example is on tuning hyperparameters in machine learning models to optimize certain model evaluation criterion; for example, the least squared error for a regression model using random forests. \cite{snoek2012practical} and \cite{7352306} frame this as a  Bayesian Optimization problem of maximizing some function on the space of hyperparameters of the class of models. Each function evaluation in this setup needs to train the model for a particular value of hyperparameter. This can be expensive when the model is large and has a large amount of training data.  The concept of reward is not meaningful in this situation because we only care about finding the model that optimizes the given loss function. Finding the best model quickly is imperative to reduce computation cost and hence it is important to know when to stop the search for the best hyperparameter value.

In order to answer this question, we would need to understand how fast $\bsx^t \rightarrow \bsx^*$ and that is the focus of this paper.  Under some assumptions, we prove concentration bounds for $\bsx^t$  where the probability that  $\bsx^t$ is bounded away from $\bsx^*$  decays exponentially fast in $t$. Moreover, the result allows us to derive an adaptive rate of convergence depending on the function structure. We show an explicit dependence on the ``sharpness" or ``flatness" of a function towards its rate of convergence. The main idea of the proof relies on breaking down the continuous domain into discrete regions and bounding the error on each discrete region, which can then be combined by the union bound. To the best of our knowledge, this is the first result that proves the convergence for an infinite-armed bandit where the utility of each arm is correlated.

The rest of the paper is organized as follows. In Section \ref{sec:problem} we formally introduce the problem, the Thompson Sampling algorithm and the main result in Theorem \ref{thm:convergence}. We describe some preliminary results in Section \ref{sec:prelims} and prove the main result in Section \ref{sec:analysis}. Simulation studies are shown in Section \ref{sec:simul_study} which highlight the convergence without the explicit assumptions required for the proof. We discuss some generalizations and concluding remarks in Section \ref{sec:conclusion}. The proofs of all preliminary and supporting results are given in the Appendix.

\section{Thompson Sampling Algorithm and The Main Result}
\label{sec:problem}

We consider the problem of sequentially maximizing a black box function $f : \cx \rightarrow \real$, where $\cx$ is a compact subset of $\real^d$. At every stage $t$ we can sample $\bsx^t \in \cx$ and observe $y^t$,  where conditionally on $\bsx^t$, $y^t$ are independent and $y^t |\bsx^t \sim N(f(\bsx^t), \sigma^2)$. 

\subsection{Gaussion Processes and Kernel Functions}
To solve the global optimization problem, we need sufficient smoothness assumptions on $f$. We assume that $f$ is a sample from a Gaussian Process (GP) with mean $0$ and kernel $k(\bsx,\bsx')$.  For any $ \bsx = (\bsx_1, \ldots, \bsx_n)$, let $\boldsymbol{f}$ denote the vectorized version of the function values obtained at the $n$ points. That is,
$\bsf = (f(\bsx_1), \ldots, f(\bsx_n))^T.$
Then, $\bsf$ is multivariate normal with mean $0$ and covariance $K$, where $K_{i,j} = k(\bsx_i,\bsx_j)$. We further assume that $k$ is a Mercer kernel on the space $\cx$ with respect to the uniform measure on $\cx$. That is, we can write $k$ as,
\begin{align*}
k(\bsx,\bsx') = \sum_{i = 1}^\infty \lambda_i \psi_i(\bsx) \psi_i(\bsx'),
\end{align*}
where $(\lambda_i)_{i \in \mathbb{N}}$ is a sequence of non-negative, non-increasing numbers, which are summable and $ (\psi_i)_{i \in \mathbb{N}}$ are a collection of mutually orthonormal functions with respect to the $L^2$ norm on $\cx$. We can consider $\lambda_i$'s to be the eigenvalues corresponding to the eigenfunctions $\psi_i$. A common example of a kernel is the Gaussian RBF-kernel, which can be parametrized by $\eta = (\ell_0, \ell_1, \ldots, \ell_d)$ and 
\begin{align*}
k_\eta(\bsx,\bsx') = \ell_0^2 \exp \left( - \frac{1}{2}\sum_{i=1}^{d}  \frac{ (\bsx_i - \bsx'_i)^2 }{\ell_i^2}  \right), 
\end{align*}
See \cite{minh2006mercer} for more details on Mercer's Theorem, kernel smoothing and many other examples.

%

\subsection{$\xi$-Greedy Thompson Sampling}

Suppose $D_t$ denotes the data we have till iteration $t-1$ and $\mathcal{F}_t$ denotes the posterior of the maximizer of $f$ given $D_t$. The  Thompson Sampling approach samples a new data point $\bsx^t$ at iteration $t$ from $\mathcal{F}_t$. We observe the data $y^{t} = f(\bsx^{t}) + \epsilon^t$, where $\epsilon^t \sim N(0,\sigma^2)$ and update $D_{t+1} = D_t \cup \{(\bsx^{t}, y^{t}) \}$. We initialize the process by assuming a non-informative prior on the distribution of the maximizer, i.e., $\mathcal{F}_0 =  U(\cx)$, the uniform distribution on $\cx$. We stop the procedure when the variance of the distribution of $\mathcal{F}_t$ becomes considerably small and we return $\bsx^* = mode(\mathcal{F}_t)$ as the estimate of the global maximizer of $f$.


In some cases, especially when $\sigma^2$ is large, this process might converge to a local optimum. Since we sample $\bsx^t$ from $\mathcal{F}_t$, we might get stuck in one place and not explore the entire space. To ensure the convergence to the global maximum, we consider an $\xi$-greedy approach. That is, with some probability $\xi > 0$, we explore the entire region $\cx$ uniformly at every stage $t$. Thus, we sample $\bsx^t \sim \mathcal{F}_t$ with probability $1-\xi$ and we sample $\bsz^t \sim U(\cx)$ with probability $\xi$. We change the notation from $\bsx^t$ to $\bsz^t$ to make it easier for the reader to differentiate between when a sample is drawn from the posterior of the maximizer $\mathcal{F}_t$ versus an uniform sample. We state the detailed steps in Algorithm \ref{algo:ts}. 


\subsection{Estimation of Hyper-Parameters}
\label{sec:hyperEst}

For simplicity of the analysis we separate the problem of estimation of hyper-parameters and learning of the function optimizer which is the main focus of this paper. In practice, we start with an initial random sample of points $D^{\textrm{random}} = \{(\bsz_1, y_1), \ldots,  (\bsz_{n_0}, y_{n_0})\}$ for some constant $n_0$. As iterations progress, whenever we 
sample $\bsz^t \sim U(\cx)$, we add it to our dataset $D^{\textrm{random}}$. Now, the estimation of the hyper-parameters is always restricted to using this random dataset $D^{\textrm{random}}$. This ensures the theoretical convergence of the $\hat{\eta} \rightarrow \eta^*$ and $\hat{\sigma} \rightarrow \sigma^*$. 

There are several methods known in literature for estimating the hyper parameters in this setup. We focus on the maximum a posteriori (MAP) estimation. For other methods see \citet{vanhatalo2012bayesian}. Here we use,
\begin{align*}
\{\hat{\eta}, \hat{\sigma}\} &= \argmax_{\eta,\sigma} p(\eta, \sigma | D^{\textrm{random}}) \\
&= \argmin_{\eta,\sigma} \left(-\log p(D^{\textrm{random}} | \eta, \sigma) - \log p(\eta,\sigma)\right),
\end{align*}
where $p(\cdot)$ denotes the likelihood function. For the Gaussian RBF kernel we can write the marginal likelihood given the parameters, $p(D^{\textrm{random}} | \eta, \sigma) = \int p(\bsy | \bsf, \sigma) p (\bsf | \bsx, \eta) \mathrm{d} \bsf$ in a closed form,
\begin{align}
\label{eq:loglikehood}
\log p(D^{\textrm{random}} | \eta, \sigma) &= C - \frac{1}{2} \log \left|K_{\eta} + \sigma^2I\right| - \frac{1}{2} \bsy^T\left(K_{\eta} + \sigma^2I\right)^{-1}\bsy,
\end{align}
where $\bsy$ denotes the vectorized version of our observed function values. Since this function is easily differentiable, we can find the optimum using any gradient descent algorithm \citep{boyd2004convex}. In situations where, a closed form expression cannot be found, we can resort to Laplace Approximations or EP's marginal likelihood approximation \citep{vanhatalo2012bayesian}. 

Note that the MAP estimator converges to the maximum likelihood estimator as we sample more and more points. Moreover, since the maximum likelihood estimator (MLE) is a consistent estimator, we assume that the regularity conditions hold such that  $\hat{\eta} \rightarrow \eta^*$ and $\hat{\sigma} \rightarrow \sigma^*$ almost surely \citep{lehmann2006theory}, where $\eta^*, \sigma^*$ denotes the true optimal parameters.


\begin{remark}
Although the usual result for consistency of the MLE only gives us convergence in probability, it is not hard to see that if we follow the proof in \citet{lehmann2006theory} we can get almost sure convergence under the extra condition that,
\begin{align}
\sup_{\theta \in \Theta} \bignorm{\hat{\ell}(\bsx | \theta) - \ell(\theta)} \xrightarrow{\text{a.s.}} 0,
\end{align}
where $\ell, \hat{\ell}$ denotes the expected log-likelihood function and its estimate, and $\theta$ is the parameter of interest.  
\end{remark}

\subsection{Sampling from the Posterior Distribution of  the maximizer}
\label{sec:max_dist}

We follow the approach in Section 2.1 of \citet{hernandez2014predictive} to sample from the distribution of the maximum given the data $D_{t}$. For sake of the proof of convergence, we choose a different feature map than what is used in \citet{hernandez2014predictive}.

Given any Mercer kernel $k_\eta$, there exists a feature map $\phi(\bsx)$ such that, $k_\eta(\bsx, \bsx') = \phi(\bsx)^T\phi(\bsx')$
where 
\begin{align*}
\phi(\bsx) = (\sqrt{\lambda_1} \psi_1(\bsx), \sqrt{\lambda_2} \psi_2(\bsx), \ldots)^T.
\end{align*}
Note that for any given $\eta_t$, we can identify the eigenvalue sequence $(\lambda_i^t)_{i \in \mathbb{N}}$. We approximate the infinite sequence by truncating the sequence at $m_t$, where $m_t$ is a sequence which is growing in the order $O(t)$. This choice of the rate of growth is a side product of the analysis and will be highlighted in the proofs of why we can use such a rate.
This enables us to approximate the kernel as
\begin{align*}
k_{\eta_t}(\bsx, \bsx') \approx \phi^t(\bsx)^T\phi^t(\bsx'),
\end{align*}
where 
\begin{align}
\label{phidef}
\phi^t(\bsx) = \left(\sqrt{\lambda_1^t} \psi_1^t(\bsx), \sqrt{\lambda_2^t} \psi_2^t(\bsx), \ldots, \sqrt{\lambda_{m_t}^t} \psi_{m_t}^t(\bsx)\right)^T.
\end{align}

Since $f$ is modeled as a sample from a Gaussian process, we can write $f(\cdot) = \phi(\cdot)^T\btheta$, where $\btheta \sim N(0, \Ib)$. Thus, to draw a sample from $\mathcal{F}_t$, we follow a two step procedure. First, we draw a random function $f^t(\cdot) = \phi^t(\cdot)^T\btheta^t$, where $\phi^t$ is given by \eqref{phidef} and $\btheta^t$ is a random vector drawn from the posterior distribution of $\btheta | (D_{t}, \phi^t)$, i.e.
\begin{equation}
\label{thetadef}
\btheta^t \sim \btheta | (D_{t}, \phi^t) = N\left(\boldsymbol{A}^{-1} \boldsymbol {\Phi}^T \boldsymbol{y}, \sigma_t^2 \boldsymbol{A}^{-1}\right),
\end{equation}
where $\boldsymbol{A} =  \boldsymbol {\Phi}^T\boldsymbol {\Phi} + \sigma_t^2 \boldsymbol{I}$ and 
\begin{align}
\label{eq:capPhi}
\boldsymbol {\Phi}^T = [\phi^t(\bsx^0), \ldots, \phi^t(\bsx^{t-1})].
\end{align}
This $f^t(\cdot)$ is an approximation to the true $f$ after observing the data $D_{t}$. Second, we generate $\bsx^t = \argmax_{\bsx \in \cx}  \phi^t(\bsx)^T \btheta^t$. This $\bsx^t$ is now a sample from $\mathcal{F}_t.$ 

\begin{remark}
Here $\bsx^0, \ldots, \bsx^{t-1}$ are the set of points in $D_t$. With a slight abuse of notation, we use $\bsx$ to denote all the points here, but $D_t$ contains samples from both $\mathcal{F}_{t}$ and $U(\cx)$ (i.e. $\bsx$ and $\bsz$).
\end{remark}  

\begin{remark}
Note that we can leverage the fact it is enough to draw samples from the posterior distribution of $f$ given the data $D_t$. We explicitly work with the feature maps $\phi^t(\cdot)$ since it makes our analysis simpler. Generating $\bsx^t = \argmax_{\bsx \in \cx}  \phi^t(\bsx)^T \btheta^t$ is exactly same as $\bsx^t = \argmax_{\bsx \in \cx} f^t(\bsx)$ where $f^t$ is drawn from the posterior of $f$ given the data $D_t$.
\end{remark}

\begin{algorithm}[!h]
\caption{$\xi$-Greedy Thompson Sampling for Infinite-Armed Bandits}\label{algo:ts}
\begin{algorithmic}[1]
\State \text{Input : Function $f$, Kernel  $k_\eta$, Domain $\cx$, Parameter $\xi, n_0$}
\State \text{Output : $\bsx^*$, the global maximum of $f$}
\State \text{Sample $\bsz_1, \ldots, \bsz_{n_0}$ uniformly from $\cx$ } 
\State Observe $y_i = f(\bsz_i) + \epsilon_i$, where $\epsilon_i \sim N(0,\sigma^2)$ for $i = 1, \ldots, n_0$.
\State \text{Set $D^{\textrm{random}} = \{(\bsz_1, y_1), \ldots,  (\bsz_{n_0}, y_{n_0})\}$}
\State Estimate the hyper-parameters $\eta_1, \sigma_1$ using $D^{\textrm{random}}$ as given in Section \ref{sec:hyperEst}.
\State \text{Set $D_1 = D^{\textrm{random}}$}
\For{$t = 1, 2, \ldots $}

\State Sample a random function $\phi$ according to \eqref{phidef} corresponding to $k_{\eta_t}$
\State Sample $\btheta^t$ from $\btheta| (D_{t}, \phi)$ according to \eqref{thetadef}
\State Compute $\bsx^t = \argmax_{\bsx \in \cx}  \phi (\bsx)^T \btheta^{t}$ and generate $\bsz^t \sim
U(\cx)$.

\State Generate a random number $\omega$.
\If {$\omega \leq \xi$}
\State Observe $y^t = f(\bsz^t) + \epsilon$.
\State $D^{\textrm{random}} = D^{\textrm{random}} \cup \{(\bsz^{t}, y^{t})\}$
\State Estimate the hyper-parameters $\eta_{t+1}, \sigma_{t+1}$ using $D^{\textrm{random}}$
\State Set $D_{t+1} =  D_{t}  \cup \{(\bsz^{t}, y^{t})\}$
\Else
\State Observe $y^t = f(\bsx^t) + \epsilon$
\State Set $\eta_{t+1} = \eta_t$ and $\sigma_{t+1} = \sigma_t$.
\State Set $D_{t+1} =  D_{t}  \cup \{(\bsx^{t}, y^{t})\}$
\EndIf
\State Break  the loop when $\bsx^t$ chosen as the maximizer converges to $\bsx^*$.
\EndFor
\State \Return $\bsx^*$
\end{algorithmic}
\end{algorithm}


\subsection{Computational Complexity}
We can assume that the function evaluation is of a constant order. Thus, if we are doing $T$ iterations, the computational complexity of generating the data is $O(T)$. Moreover, the estimation of the hyper-parameters is done using gradient descent whose computational complexity is $O(\log(1/\kappa))$ to get an accuracy of $\kappa$ for the objective function. Thus, for $T$ iterations in expectation we need to run $O(\xi T \log(1/\kappa))$ iterations of the gradient descent. Our eigenfunction corresponding to the kernel would be known and hence function evaluation is again of constant order. Our major computation comes the matrix inversion that is necessary in \eqref{eq:loglikehood} and \eqref{thetadef}. At the $t$-th iteration we would need $O(t^3)$ operations. Thus, in the worst case if we run it for $T$ iterations we would need $O(T^4)$ operations. This is the same worst-case computational complexity for running any similar GP-UCB type algorithms as well \citep{Rasmussen:2005:GPM:1162254}.  

\subsection{Main Result}
\label{sec:reg_cond}
The main aim of the paper is to prove a concentration bound for $\bsx^t$ for the Thompson Sampling approach from Algorithm \ref{algo:ts}. We need some further assumptions and regularity conditions to achieve that.

\begin{assumption}
\label{assu:func_norm} 
Let $\phi^{t}$ be the feature map for $k_{\eta_t}$. Then, there exists a sequence of $\btheta_{t}^*$  with $\norm{\btheta_t^*} \leq \sqrt{t}M$ such that,
$$
\limsup_{t \rightarrow \infty} \sup_{\bsx \in \cx} \left| f(\bsx) - \phi^{t}(\bsx)^T\btheta_{t}^*\right|  = 0
$$
almost surely. Here $M$ is a positive constant and the function $f$ is assumed to be coming from a Gaussian Process with kernel parameters $\eta^*$. 
\end{assumption}

The above assumption says that outside of a measure zero set, for large enough $t$ we have, $$\sup_{\bsx \in \cx} \left| f(\bsx) - \phi^{t}(\bsx)^T\btheta_{t}^*\right|  < \delta_0(t),$$ where $\delta_0(t)$ is a decreasing function of $t$ converging to 0. The exact rate of decay depends on $f$, but throughout the proof we only require that $\delta_0(t)$ is a decreasing function. Note that since $f$ is assumed to come from a Gaussian Process, it can be written as a linear combination of the feature maps from the kernel. Now, since kernel hyperparameters $\eta_t$ converges to the truth $\eta^*$, this assumption intuitively holds. 
\begin{assumption}
\label{assu:eigen_func}
The kernel must belong to either of the following two classes. 
\begin{itemize}
\item[(a)] \textit{Bounded eigen functions.} For this class of kernels, there exists an $M$ such that $$|\psi_i(\bsx)| \leq M\;\; \text{ for all } i.$$ An example of this is when $\psi_i$ form a sine basis on $\cx = [0, 2\pi]$ \citep{braun2006accurate}.
\item[(b)] \textit{Bounded kernel functions.} For this class of kernels, there exists an $M$ such that 
$$
k(\bsx, \bsx) \leq M < \infty\;\;\; \text{ for all } \bsx \in \cx.
$$
A very typical example in this class is the RBF kernel, or the squared exponential kernel. All shift-invariant kernels fall in this category.
\end{itemize}
\end{assumption}


\begin{assumption}
\label{assu:lip}
There exists a $C$ such that for all $\bsx, \bsy \in \cx$, 
$$
k(\bsx, \bsx) - k(\bsx, \bsy) \leq C\norm{\bsx - \bsy}^2. 
$$
\end{assumption}

Most common kernel satisfy this constraint. For example, considering the RBF kernel $k(\bsx,\bsy) = \exp(-\norm{\bsx - \bsy}/2\ell^2)$ we have,
\begin{align}
\label{eq:rbf_lip}
k(\bsx, \bsx) - k(\bsx, \bsy) = 1 - \exp\left(-\frac{\norm{\bsx - \bsy}^2}{2\ell^2}\right) \leq C\norm{\bsx - \bsy}^2.
\end{align}
For a thorough list and more examples see \cite{minh2006mercer}.   

Let us introduce one more notation that we use throughout the rest of the paper. Let us define $\delta_\epsilon$ as the minimum difference in the function values between the optimal $\bsx^*$ and any $\bsx$ which is at least $\epsilon$ distance away from the optimal. Formally,
\begin{align}
\label{eq:deltae}
\delta_\epsilon := \inf_{\bsx: \bignorm{\bsx - \bsx^*} > \epsilon} f(\bsx^*) - f(\bsx).
\end{align}
We know that $\delta_\epsilon>0$ since $f$ has an unique maximum. $\delta_\epsilon$ measures a degree of sharpness of the function around its true global maximum. The adaptive convergence rate as discussed in this paper is formalized through this parameter. With these assumptions and notations we can now state our main result.

\begin{theorem}
\label{thm:convergence} 
Let $f$ be a sample from a Gaussian Process on a compact set $\cx \subset \real^d$ having a global unique maximum at $\bsx^*$. Then, under Assumptions \ref{assu:func_norm} - \ref{assu:lip}, if we follow the Thompson Sampling procedure as given in Algorithm \ref{algo:ts}, there exists a $T$ such that for all $t > T$, 
$$ P(\norm{\bsx^t - \bsx^*} > \epsilon) \leq  C \frac{t^{d/2}}{\delta_\epsilon^d} \exp(-c \delta_\epsilon^2 t),$$
where $C, c$ are positive constants and $\delta_\epsilon$ is defined in \eqref{eq:deltae}.
\end{theorem}

\subsection{Discussion}
\label{sec:discThm}
Theorem \ref{thm:convergence} gives us an explicit rate of decay of the distance between $\bsx^t$ and $\bsx^*$. Note that the explicit rate of convergence is adaptive as it depends on how sharp or flat the function $f$ truly is, as formalized through $\delta_\epsilon$. 

Intuitively, if the function is very flat, for example, the Gaussian density with an extremely large variance, then $\delta_\epsilon$ would be quite small and hence convergence of $\bsx^t$ to $\bsx^*$ would be very slow.  On the contrary, if the function has a sharp peak, $\delta_\epsilon$ is large and hence the covergence will be much faster. 

As a concrete example, if $f(x) = -|x|$, then $\delta_\epsilon = \epsilon$, in which case, we can actually derive the explicit rate of convergence. Specifically we get,
$$ P(\norm{\bsx^t - \bsx^*} > \epsilon) \leq  C \frac{\sqrt{t}}{\epsilon^2} \exp(-c \epsilon^2 t).$$ Thus, the rate of convergence of $\bsx^t$ to $\bsx^*$ is $O(t^{-1/2 + \rho})$ where $\rho > 0$ is arbitrary. Hence, for any function with a sharper peak around its global maximum the rate of convergence is faster, while for a function which is more flat, the rate of convergence is slower. 

Overall, the function structure plays an important role in determining the explicit rate of convergence. We show a simple simulation example in Section \ref{sec:simul_study} to see this decay rate as a function degenerates into a flat function.

\section{Preliminaries}
\label{sec:prelims}

We now state some preliminary results, which will be used throughout the rest of the proof. For the rest of the paper, we denote the changing constant as $c$. Also, throughout the paper we make statements  for Algorithm \ref{algo:ts} under Assumptions \ref{assu:func_norm} - \ref{assu:lip} without explicitly stating it every time. The first result gives a bound on the minimum and maximum eigenvalues of the matrix $\boldsymbol{A}/t$. 

\begin{lemma}
\label{lemma:eigen}
Let $\boldsymbol{A} =  \boldsymbol {\Phi}^T\boldsymbol {\Phi} + \sigma_t^2 \boldsymbol{I}$, where $\boldsymbol {\Phi}$ is defined \eqref{eq:capPhi}. Then, 
\begin{align*}
\liminf_{t \rightarrow \infty} \lambda_{\min} \left(\frac{\boldsymbol{A}}{t}\right) \geq \xi c > 0 \;\;\; \text{ a. s.,}
\end{align*}
and
\begin{align*}
\limsup_{t \rightarrow \infty} \lambda_{\max} \left(\frac{\boldsymbol{A}}{t}\right) \leq \xi C  + \alpha (1 - \xi) + 1 < \infty  \;\;\; \text{ a. s.,}
\end{align*}
where $\alpha = k_{\eta}(\bsx, \bsx)$ and $C, c$ are constants.
\end{lemma}

\begin{remark}
Note that, Lemma \ref{lemma:eigen} and Assumption \ref{assu:func_norm} make almost sure statements. If we denote our probability space by $(\Omega, \Fcal, \Pcal)$, then, throughout the rest of this paper, we only work over those set of $\omega \in \Omega$ where the statements in Lemma \ref{lemma:eigen} and Assumption \ref{assu:func_norm} hold. Moreover, we use limsup and liminf since the limit may not exist.
\end{remark}

As a corollary to Lemma \ref{lemma:eigen} we can show upper bounds to much more complicated matrix forms involving $\boldsymbol{A}$. Two such results, which will be used later are as follows.
\begin{lemma}
\label{lemma:prod_eigen}
There exists a constant $c > 0$ such that for all large enough $t$,
$$
\lambda_{\max} \left(  \left(\frac{\boldsymbol{A}}{t}\right)^{-1} \frac{\boldsymbol {\Phi}^T\boldsymbol {\Phi}}{t} \left(\frac{\boldsymbol{A}}{t}\right)^{-1} \right) \leq c.
$$
\end{lemma}

\begin{lemma}
\label{lemma:main_assumption}
The following bounds hold: 
\begin{itemize}
\item[(a)] $\bignorm{\boldsymbol{A}^{-1} \boldsymbol {\Phi}^T  \boldsymbol {\Phi} - \boldsymbol{I}} \leq c/t,$
\item[(b)] $ \bignorm{\boldsymbol{A}^{-1} \boldsymbol {\Phi}^T} \leq C/\sqrt{t} ,$
\end{itemize}
where $c, C$ are constants and $\bignorm{\cdot}$ denotes the spectral norm of the matrix.
\end{lemma}

The following result quantifies that if $f$ can be approximated well, then there is a positive difference between $f(\bsx^*)$ and $f(\bsx)$ for any $\bsx$ which is at least $\epsilon$ distance away from the optimal $\bsx^*$. Formally, we show the following.

\begin{lemma}
\label{lemma:numerator_bound}
Given $\epsilon > 0$, for any $\bsx$ such that $\bignorm{\bsx - \bsx^*} > \epsilon$ and large enough $t$,
\begin{align*}
\left(\phi^t(\bsx^*) - \phi^t(\bsx)\right)^T \boldsymbol{A}^{-1} \boldsymbol {\Phi}^T \bsf \geq \frac{\delta_\epsilon}{2} > 0.
\end{align*} 
\end{lemma}

Our last preliminary result shows a concentration bound on Chi-square random variables, which will be needed for subsequent proofs.

\begin{lemma}
\label{lemma:chi_exp}
Let $Z \sim \chi_m^2$. Then for any $\delta > 0$,
$$P( Z \geq m + \delta) \leq \exp\left(-\frac{1}{2} \left(\delta + m + \sqrt{2\delta m + m^2}\right)\right).$$
\end{lemma}

We set one last notation that we use in the proofs below. By Lemma \ref{lemma:eigen} and Lemma \ref{lemma:prod_eigen}, for large enough $t$ we have,
\begin{equation}
\begin{aligned}
\label{eq:constant_def1}
&\lambda_{\min}\left(\frac{\boldsymbol{A}}{t}\right) \geq c_1, \\
&\lambda_{\max} \left(  \left(\frac{\boldsymbol{A}}{t}\right)^{-1} \frac{\boldsymbol {\Phi}^T\boldsymbol {\Phi}}{t} \left(\frac{\boldsymbol{A}}{t}\right)^{-1} \right) \leq c_2,
\end{aligned}
\end{equation}
where $c_1$ and $c_2$ are constants. Furthermore, using Lemma \ref{lemma:main_assumption} and Assumption \ref{assu:func_norm} we can write, 
\begin{align}
\label{eq:constant_def2}
\norm{\boldsymbol{A}^{-1} \boldsymbol {\Phi}^T \bsf} &= \norm{\boldsymbol{A}^{-1} \boldsymbol {\Phi}^T \boldsymbol {\Phi} \btheta^* - \delta_0(t) \boldsymbol{A}^{-1} \boldsymbol {\Phi}^T\one}  \nonumber\\
&= \norm{\btheta^* + \boldsymbol{E}\btheta^* - \delta_0(t) \boldsymbol{A}^{-1} \boldsymbol {\Phi}^T\one} \nonumber \\
&\leq \norm{\btheta^*} + \norm{\boldsymbol{E}\btheta^*} + \norm{\delta_0(t) \boldsymbol{A}^{-1} \boldsymbol {\Phi}^T\one} \nonumber \\
&\leq \norm{\btheta^*}\left(1 + \norm{\boldsymbol{E}}\right) + \delta_0(t) \sqrt{t} \norm{\boldsymbol{A}^{-1} \boldsymbol {\Phi}^T} \nonumber \\
&\leq \sqrt{c_3 t},
\end{align}
where $\boldsymbol{E} = \boldsymbol{A}^{-1} \boldsymbol {\Phi}^T \boldsymbol {\Phi} - \boldsymbol{I}$ and $\delta_0(t)$ is a decreasing function of $t$.

\section{Proof of Theorem \ref{thm:convergence}}
\label{sec:analysis}

\subsection{Outline}
The main idea of the proof relies on breaking down the continuous domain into small regions and bounding the errors on each region, which are then combined using the union bound. To bound the error on each small region, we compare the function values at a single point within the region and bound the error appropriately. In order to first compare the function values at a single point, we rely on the following Lemma.

\begin{lemma}
\label{lemma:prob_bound}
For any $\bsx$ such that $\bignorm{\bsx - \bsx^*} > \epsilon$, any $0 < \epsilon' \leq  \delta_\epsilon/4$, for $t$ large enough
$$P \left( \phi^t(\bsx^*)^T\btheta^t < \phi^t(\bsx)^T\btheta^t  + \epsilon' \right) \leq
2\exp(-c \delta_\epsilon^2 t), $$
where $\delta_\epsilon$ is defined as in \eqref{eq:deltae} and $c$ is a positive constant. 
\end{lemma}

Once we have this control, in order to bound the supremum in the sub-region, we a use a truncation argument. We truncate on $\bignorm{\btheta^t}$ and bound the probability of  $\bignorm{\btheta^t} $ exceeding the truncation value. This is done, using the following result.

\begin{lemma}
\label{lemma:norm_bound}
For any $\delta > 0$ and $t$ large enough, let
$$
L_t :=   \sqrt{\sigma_t^2 \left( \frac{m_t}{t} + \delta\right)\left(\frac{1}{c_1} + \frac{1}{c_2} \right) + c_3 t},
$$
where $c_1, c_2, c_3$ are defined in \eqref{eq:constant_def1} and \eqref{eq:constant_def2}. Then,
\begin{align*}
P&\left(\bignorm{\btheta^t} > L_t\right) \leq 2\exp\left(-\frac{\delta t}{2}\right).
\end{align*}
\end{lemma}

Finally, we show that we can appropriately choose the number of discrete sub-regions such that the union bound converges, which will be enough for the proof. The details are now given below.
\subsection{Proof of Theorem \ref{thm:convergence}}
We begin by observing that for any $\epsilon > 0$,
\begin{align*}
&P(\bignorm{\bsx^t - \bsx^*} > \epsilon) \\
&= P\left(\sup_{\bsx \in \mathcal{B}_{\epsilon}(\bsx^*)} \phi^t(\bsx)^T\btheta^t < \sup_{\bsx \in \cx \setminus \mathcal{B}_{\epsilon}(\bsx^*)} \phi^t(\bsx)^T\btheta^t\right) \\
&\leq P\left( \phi^t(\bsx^*)^T\btheta^t < \sup_{\bsx \in \cx \setminus \mathcal{B}_{\epsilon}(\bsx^*)} \phi^t(\bsx)^T\btheta^t\right),
\end{align*}
where $\mathcal{B}_{\epsilon}(\bsx^*)$ denotes an open ball of radius $\epsilon$ around $\bsx^*$.
Now since $\cx \setminus \mathcal{B}_{\epsilon}(\bsx^*)$ is a compact set, and a metric space with respect to the Euclidean norm, we can cover it with an $\epsilon$-Net \citep{vershynin2010introduction}. Specifically, for any $\epsilon_t > 0$, there exists a finite subset $\Ncal_{\epsilon_t}$ of $\cx \setminus \mathcal{B}_{\epsilon}(\bsx^*)$ such that given any $\bsx \in \cx \setminus \mathcal{B}_{\epsilon}(\bsx^*)$, there exists a $\bsy \in \Ncal_{\epsilon_t}$ such that $\bignorm{\bsx - \bsy} < \epsilon_t$. Moreover, $\Ncal_{\epsilon_t}$ can be chosen such that,
$$
\left| \Ncal_{\epsilon_t} \right| \leq \left(\frac{3}{\epsilon_t}\right)^d \frac{\textrm{vol}(\cx \setminus \mathcal{B}_{\epsilon}(\bsx^*))}{\textrm{vol}(B)},
$$
where $\textrm{vol}(\cdot)$ denotes the volume and $B$ denotes the unit ball in dimension $d$ (See Lemma 5.2 of \cite{vershynin2010introduction} for a detailed proof). For simplicity, let $\{\bsx_i\}_{i=1}^{\Ncal_{\epsilon_t}}$ denotes the set of points in the $\epsilon_t$-Net. Thus, we can write
\begin{align*}
P(\bignorm{\bsx^t - \bsx^*} > \epsilon) \leq   P\left( \phi^t(\bsx^*)^T\btheta^t < \max_{i = 1,\ldots, \Ncal_{\epsilon_t}} \sup_{\bsx \in \mathcal{B}_{\epsilon_t}(\bsx_i)} \phi^t(\bsx)^T\btheta^t\right).
\end{align*}
Now for any $\bsx \in \mathcal{B}_{\epsilon_t}(\bsx_i)$,  we have
\begin{align*}
\bignorm{\phi^t(\bsx) - \phi^t(\bsx_i)}^2 &=  \sum_{j=1}^{m_t} \lambda_j^t \left[ \psi_j^t(\bsx)^2 + \psi_j^t(\bsx_i)^2 - 2 \psi_j^t(\bsx) \psi_j^t(\bsx_i)\right] \\
&= \sum_{j=1}^{m_t} \lambda_j^t \left( \psi_j^t(\bsx) - \psi_j^t(\bsx_i)\right)^2 \\
&\leq \sum_{j=1}^{\infty} \lambda_j^t \left( \psi_j^t(\bsx) - \psi_j^t(\bsx_i)\right)^2 = k_{\eta_t}(\bsx, \bsx) + k_{\eta_t}(\bsx_i, \bsx_i) - 2k_{\eta_t}(\bsx, \bsx_i) \\
&\leq  2C \norm{\bsx - \bsx_i}^2 = 2C \epsilon_t^2 
\end{align*}
where last inequality follows from Assumption \ref{assu:lip}. Thus, for any $\delta > 0$, from Lemma \ref{lemma:norm_bound} choosing 
\begin{align*}
 L_t &= \sqrt{\sigma_t^2 \left( \frac{m_t}{t} + \delta\right)\left(\frac{1}{c_1} + \frac{1}{c_2} \right) + c_3 t}  = O\left(\sqrt{t}\right),
\end{align*}
and 
$\bignorm{\btheta^t} \leq L_t$, we get,
\begin{align*}
\bignorm{(\phi^t(\bsx) - \phi^t(\bsx_i))^T\btheta^t} &\leq \bignorm{\btheta^t} \bignorm{\phi^t(\bsx) - \phi^t(\bsx_i)} \leq C L_t \epsilon_t.
\end{align*}
Thus conditioning on $\bignorm{\btheta^t} \leq L_t$, we get,  
\begin{align*}
 \sup_{\bsx \in \mathcal{B}_{\epsilon_t}(\bsx_i)} \phi^t(\bsx)^T\btheta^t \leq \phi^t(\bsx_i)^T\btheta^t  +  CL_t \epsilon_t.
\end{align*}
Using this we get,
\begin{align*}
&P(\bignorm{\bsx^t - \bsx^*} > \epsilon) \\&\leq   P\left( \phi^t(\bsx^*)^T\btheta^t < \max_{i = 1,\ldots, \Ncal_{\epsilon_t}} \phi^t(\bsx_i)^T\btheta^t +  CL_t \epsilon_t \right) + P(\bignorm{\btheta^t} > L_t) \\
&\leq \sum_{i=1}^{\Ncal_{\epsilon_t}} P\left( \phi^t(\bsx^*)^T\btheta^t < \phi^t(\bsx_i)^T\btheta^t +  CL_t \epsilon_t \right)  + P(\bignorm{\btheta^t} > L_t) \\
&\leq 2 \left\lvert\Ncal_{\epsilon_t}\right\rvert \exp(-c \delta_\epsilon^2 t) + 2\exp\left(-\frac{\delta t}{2}\right),
\end{align*}
where the first inequality follows by conditioning on $\bignorm{\btheta^t} \leq L_t$, the second from the union bound and the last inequality by using Lemmas \ref{lemma:prob_bound} and \ref{lemma:norm_bound}. 

To satisfy Lemma \ref{lemma:prob_bound} we choose $\epsilon_t = \delta_\epsilon/16CL_t.$ 
This gives us,
\begin{align*}
 \left\lvert\Ncal_{\epsilon_t}\right\rvert \leq \left(\frac{3}{\epsilon_t}\right)^d\frac{\textrm{vol}(\cx \setminus \mathcal{B}_{\epsilon}(\bsx^*))}{\textrm{vol}(B)} =  c\left(\frac{48 C L_t}{\delta_\epsilon}\right)^d = O\left(t^{\frac{d}{2}}\right).
\end{align*}
Now, since $d$ is finite, $\left\lvert\Ncal_{\epsilon_t}\right\rvert$ grows to infinity much slower than exponential. Hence, we get for some positive constants $C, c$ and $t$ large enough,
\begin{align*}
P(\bignorm{\bsx^t - \bsx^*} > \epsilon) &\leq C \frac{t^{d/2}}{\delta_\epsilon^d} \exp(-c \delta_\epsilon^2 t) + 2\exp\left(-\frac{\delta t}{2}\right).
\end{align*}
Finally, since the choice of $\delta$ was arbitrary, we have,
\begin{align*}
P(\bignorm{\bsx^t - \bsx^*} > \epsilon) &\leq C \frac{t^{d/2}}{\delta_\epsilon^d} \exp(-c \delta_\epsilon^2 t). 
\end{align*}
This concludes the proof of Theorem \ref{thm:convergence}.

\section{Simulation Study}
\label{sec:simul_study}
We have proved the  convergence for the $\xi$-greedy Thompson Sampling algorithm, under  Assumptions \ref{assu:func_norm} - \ref{assu:lip}. We know that the algorithm converges asymptotically. Through simulations, we first show how it converges in practice with different amount of noise and different dimensions. We also show how the structure of the function affects the rate of convergence as explained in Section \ref{sec:discThm}. 

\subsection{Convergence in Practice}
 Throughout this study, we have considered the kernel to be the RBF-kernel, which satisfies all the regularity conditions. Specifically for $\eta = (\ell_0, \ell_1)$ our kernel function is defined as
 \begin{align*}
 k_\eta(\bsx, \bsx') = \ell_0^2 \exp\left(- \frac{\norm{\bsx - \bsx'}^2}{2\ell_1^2}\right).
 \end{align*}
This kernel satisfies all the assumptions as mentioned in Section \ref{sec:reg_cond}. We further choose $n_0 = 30$, $\xi = 0.15$ and the  truncation parameter $m_t = 1000t$ in our simulations. 

Overall, we take a batch approach during the iteration of the algorithm. At each stage instead of drawing a single point, we draw $30$ points from the distribution of the maximum. Specifically, with probability $\xi$ we generate $\bsz^{t_1}, \ldots,  \bsz^{t_{30}} \sim U(\cx)$ and with probability $1 - \xi$ we generate $$\bsx^{t_j} = \argmax_{x \in \cx} \phi(\bsx)^T \btheta^{t_j},$$ for $j = 1, \ldots, 30$. This is done to increase the exploration part of the algorithm while keeping the running time constant. 

We describe two simulation results, one with a 1-dimensional function and another with a bivariate function.  For the 1-dimensional example, we consider a bimodal function 
\begin{align}\label{eq:f1}
f^1(x) &:= \frac{5}{\sqrt{2\pi}}\exp\left(-\frac{(x - 2)^2}{2}\right)  + \frac{10}{\sqrt{2\pi}}\exp\left(-\frac{(x - 5)^2}{2}\right),
\end{align}
which has a local maximum at $2$ and a global maximum at $5$. We consider a similar function in a 2-dimensional space
\begin{align}\label{eq:f2}
f^2(x) &:= \frac{5}{2\pi}\exp\left(-\frac{1}{2}\bignorm{\bsx - \mu_1}^2\right)  + \frac{10}{2\pi}\exp\left(-\frac{1}{2}\bignorm{\bsx - \mu_2}^2\right),
\end{align}
where $\mu_1 = (2, 2)$ is the local maximum and $\mu_2 = (5, 5)$ is the global maximum. At every trial, we can draw a single value of $x$ and observe $y=f(x) + \epsilon$, where $\epsilon$ are i.i.d. Gaussian random errors with mean $0$ and standard deviation $\sigma \in [0.1, 5]$. 

\begin{figure}[!h]
\centering
\begin{subfigure}{.48\textwidth}
  \centering
  \includegraphics[width=\linewidth]{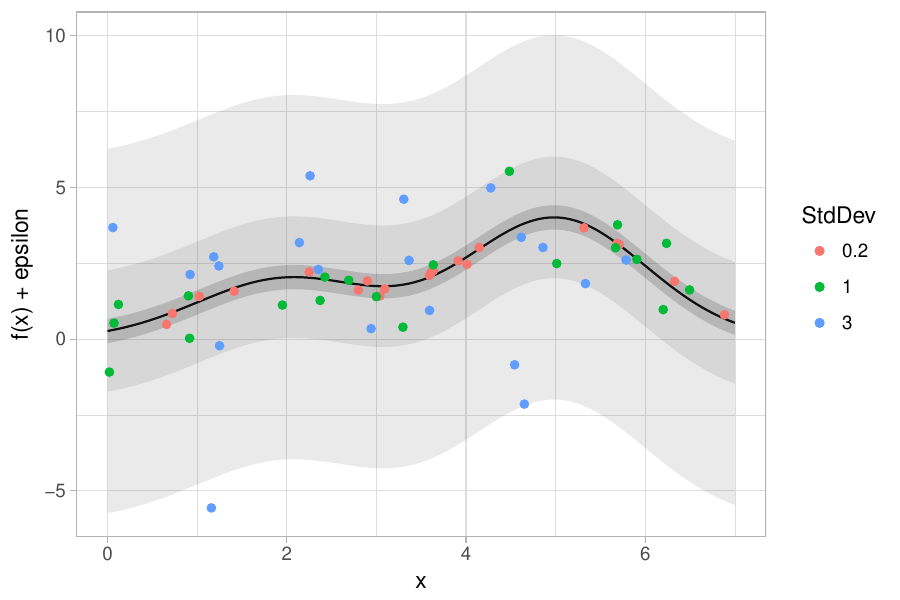}
  \caption{$f^1(x)$, 1-dim function in \eqref{eq:f1} with some realizations}
  \label{fig:sub1}
\end{subfigure}%
\begin{subfigure}{.48\textwidth}
  \centering
  \includegraphics[width=\linewidth]{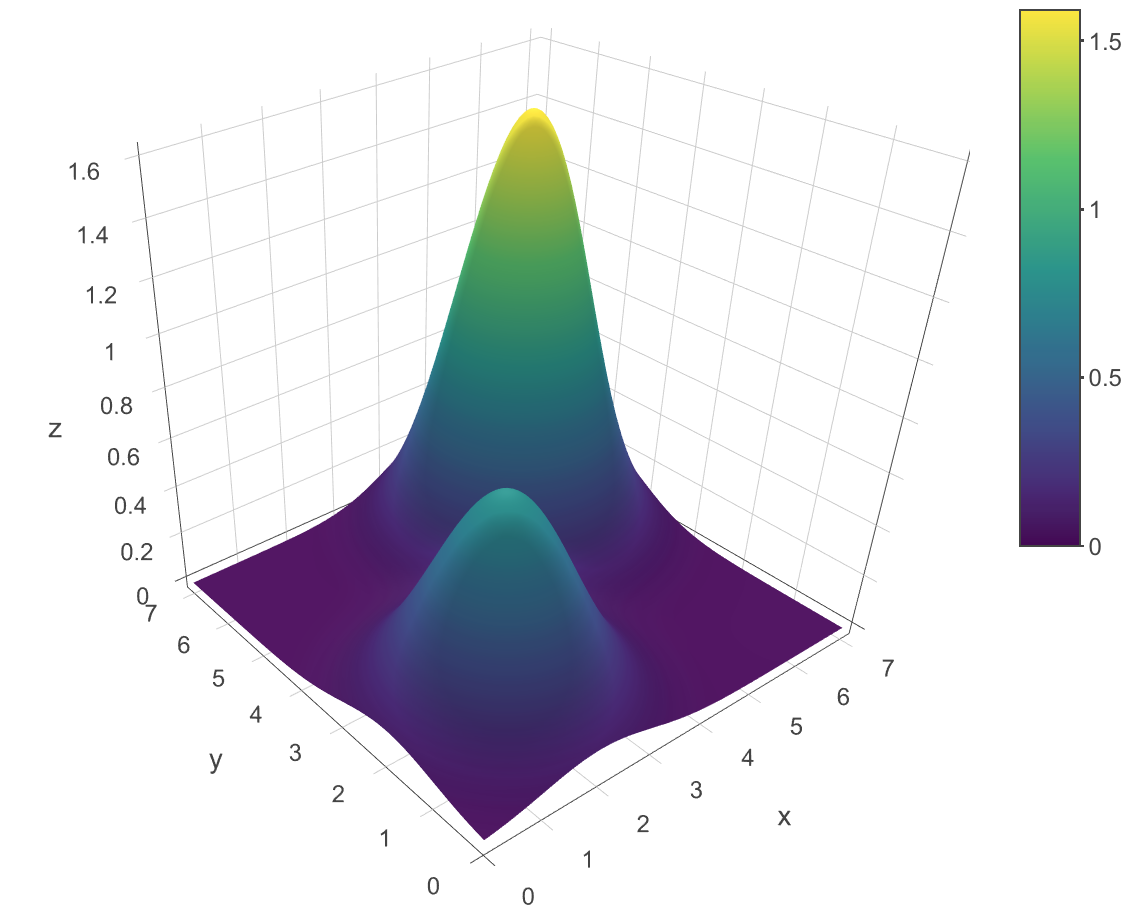}
  \caption{$f^2(x)$, 2-dim function in \eqref{eq:f2}}
  \label{fig:sub2}
\end{subfigure}
\caption{Plot of functions described in \eqref{eq:f1} and \eqref{eq:f2}.}
\label{fig:funcplots}
\end{figure}

Figure \ref{fig:funcplots}(a) shows the univariate function, along with the 95\% confidence bands for the sampling error. It also shows a sample of points obtained when using different standard deviations in the error generation mechanism. Note that, as the error standard deviation $\sigma$ increases, it becomes increasingly hard to identify the true function.  Figure \ref{fig:funcplots}(b) shows the bivariate function $f^2(\bsx)$. Although we do not add the confidence bands and sample points in the figure because it is hard to visualize in a plot, we do work with a wide range of the error.

\begin{figure}[H]
\centering
\begin{subfigure}{.47\textwidth}
  \centering
  \includegraphics[width=\linewidth]{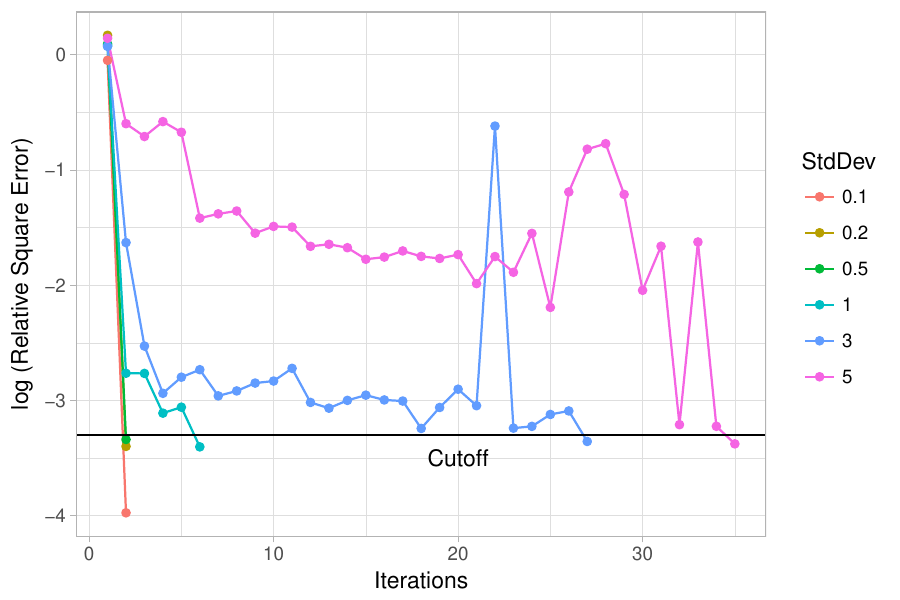}
  \caption{$f^1(x)$ defined in \eqref{eq:f1} }
\end{subfigure}%
\begin{subfigure}{.47\textwidth}
  \centering
  \includegraphics[width=\linewidth]{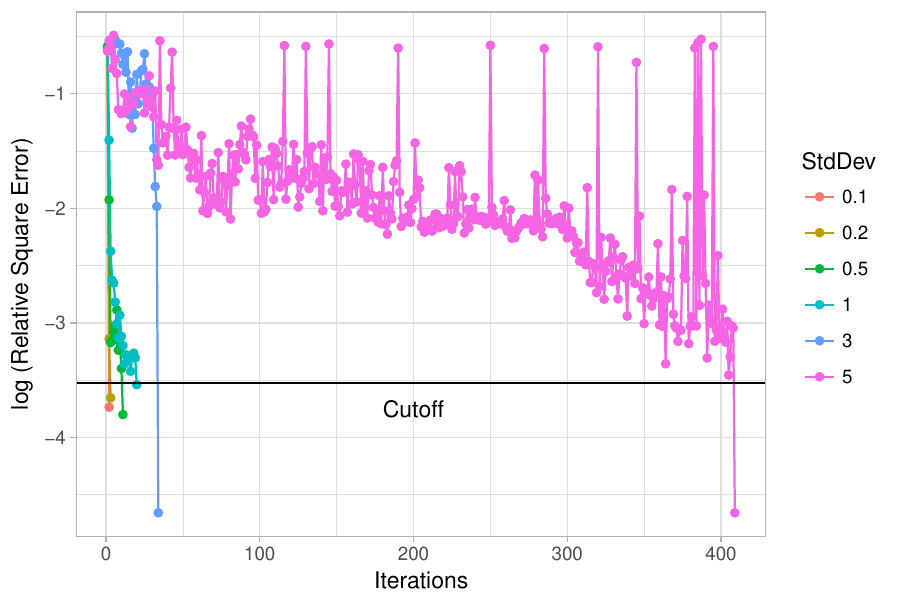}
  \caption{$f^2(x)$ defined in \eqref{eq:f2}}
\end{subfigure}
\caption{Decay rate for different values of $\sigma$.}
\label{fig:error_sub}
\end{figure}

Figure \ref{fig:error_sub}(a) and \ref{fig:error_sub}(b) show the decays in the relative squared error for different values of standard deviation $\sigma$. We plot log of the relative error, $2\log_{10}(( \bsx_t - \bsx^*)/\bsx^*)$ vs iteration $t$. For smaller $\sigma$, we see very quick convergence for both of the example functions. As the errors increase we see that the algorithm takes a longer time to converge. The sudden spikes in the error plot are because of the iterations where we do random sampling instead of sampling from the maximum. Moreover, we notice that in general, the number of iterations required to converge for a 2-dimensional function is larger than that for a 1-dimensional case, especially when the errors increase in the observations as shown through the dependence on $d$ in Theorem \ref{thm:convergence}.

\begin{figure}[H]
\centering
\includegraphics[width=0.5\linewidth]{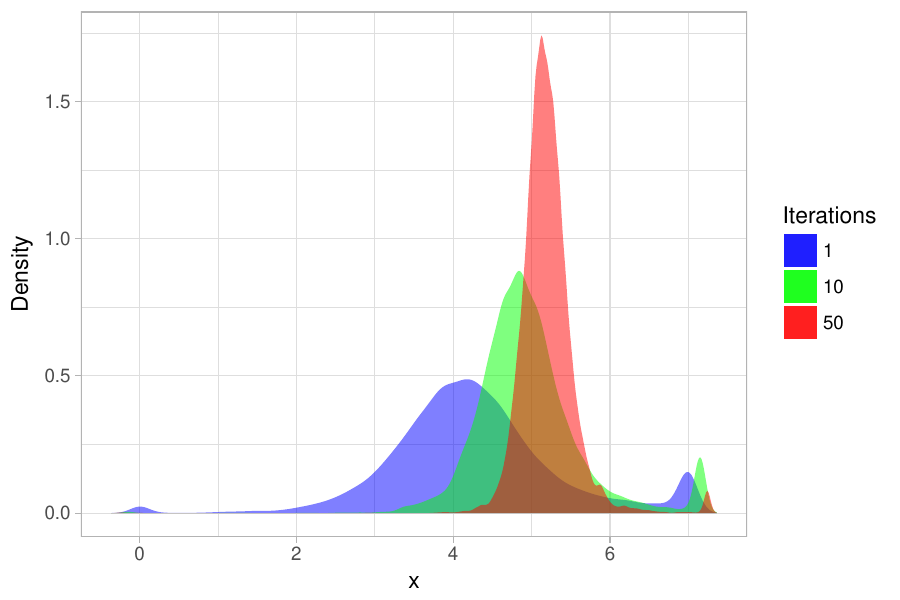}
  \caption{Sequence of estimated distributions of the maximum across iterations for $\sigma = 5$ when trying to estimate the maximum of $f^1(\bsx)$. A similar figure is seen for the bivariate function, $f^2(\bsx)$ as well.}
 \label{fig:1dalgo}
\end{figure}

Figure \ref{fig:1dalgo}  shows the distribution of the maximum across different iterations when we have $\sigma = 5$. We see that as the iterations increase, the distribution concentrates around $\bsx^*$. At each stage, sampling from the probability measure brings us closer to the true maximum while allowing some room to explore.

These examples show that without explicit assumptions on the function we can converge to the true global maximum even when there is a large level of noise in the data. 

\subsection{Function Structure} 
\begin{figure}[H]
\centering
\begin{subfigure}{.5\textwidth}
  \centering
  \includegraphics[width=\linewidth]{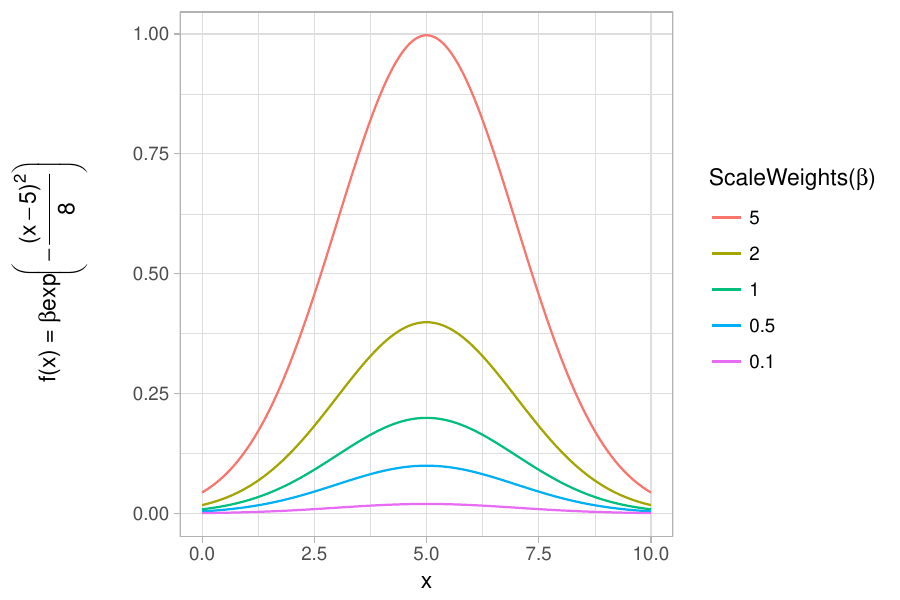}
  \caption{The function $f_\beta(x)$  }
\end{subfigure}%
\begin{subfigure}{.5\textwidth}
  \centering
  \includegraphics[width=\linewidth]{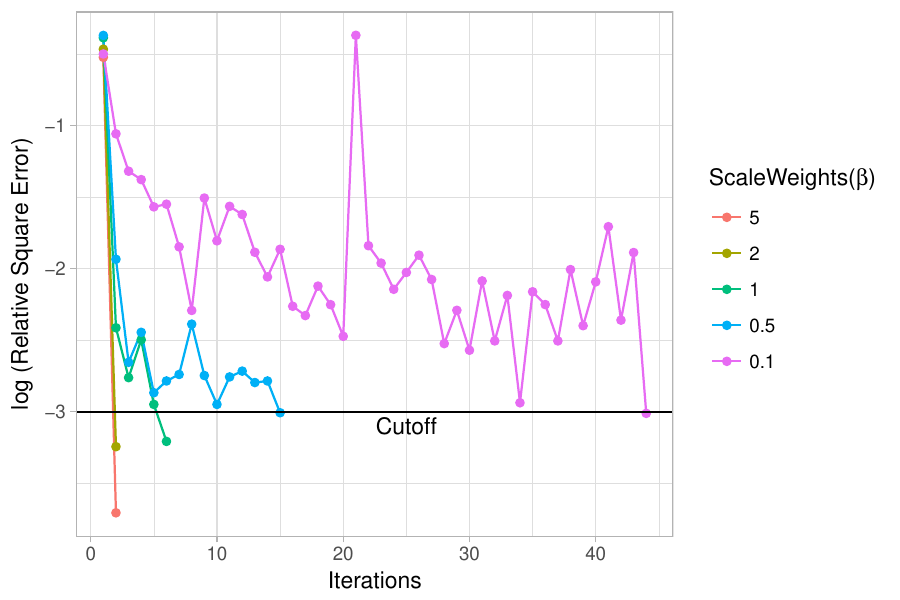}
  \caption{Error decay for function $f_\beta(x)$}
\end{subfigure}
\caption{Simulations for the function $f_\beta(x)$ for different values of $\beta$. We see that finding the maximum is easier when it is more prominent.}
\label{fig:f_beta}
\end{figure}

Although we have seen convergence, it is a difficult problem to characterize how quickly we start to see that rate of decay, since it depends on the structure of the underlying function as shown through Theorem \ref{thm:convergence}. To observe how the decay rate actually changes in practice, we consider a simple example with the following one-dimensional function,
\begin{align*}
f_\beta(x) = \beta \exp\left(-\frac{(x-5)^2}{8}\right).
\end{align*}
This is a unimodal function with a global maximum at $5$ and as we decrease the value of $\beta$ the function becomes flatter ($\delta_{\epsilon}$ decreases). Figure \ref{fig:f_beta}(a) shows this behavior across different value of $\beta$. While trying to estimate the maximum of each of the above functions, we keep a constant error rate of $\sigma = 0.1$. In this setup, we observe the decay rate as shown in Figure \ref{fig:f_beta}(b). For a fixed error rate $\sigma$, the rate of convergence slows down as the function becomes flatter as expected through Theorem \ref{thm:convergence}.

\section{Conclusion}
\label{sec:conclusion}
We have proved an adaptive rate of convergence for the $\xi$-Greedy Thompson Sampling Algorithm in the case of an infinite armed bandit with a Gaussian Process prior on the reward.  The rate is formalizes through a sharpness parameter $\delta_\epsilon$. As far as we know, this is the first result which quantifies the rate of decay of the sequential point $\bsx^t$ to the true optimal $\bsx^*$. Although we have proved the result where at every stage we are sampling only one point, it should be easy to generalize to more points and we leave it as a future work. While actually running the algorithm in practice, we use the batch mode to sample more points at every stage as done throughout our simulations in the Section \ref{sec:simul_study}. By doing so, we can explore the function better in parallel within the same running time as a single point evaluation. The simulation study under this regime shows quick convergence as well as the fact that the assumptions are not too restrictive. This novel proof technique that we have presented here, can be now used to solve a variety of problems and opens up a new direction of research. Using this technique explicit (adaptive) convergence rates can be shown for most of the UCB type algorithms. We leave such generalizations as future work.

\section*{Acknowledgment}
We would like to thank Deepak Agarwal, Liang Zhang, Yang Yang, Ying Xuan, Preetam Nandy and Rajarshi Mukherjee for the several fruitful discussions regarding this paper.
\bibliography{oms}

\section{Appendix}
\label{sec:proofs}
We collect the proofs of all the preliminary and supporting Lemmas here.

\subsection*{Proof of Lemma \ref{lemma:eigen}}
We begin with the lower bound. Observe that,
\begin{align*}
\lambda_{\min} \left(\frac{\boldsymbol{A}}{t}\right) &= \lambda_{\min} \left(\frac{\boldsymbol {\Phi}^T\boldsymbol {\Phi}}{t} + \frac{ \sigma_t^2 \boldsymbol{I}}{t}\right) \\
& \geq \lambda_{\min} \left(\frac{\boldsymbol {\Phi}^T\boldsymbol {\Phi}}{t} \right) = \lambda_{\min} \left(\frac{1}{t}\sum_{i = 0}^{t-1} \phi^t(\bsx^i) \phi^t(\bsx^i)^T\right).
\end{align*}
We separate out the sum into cases where the samples are from $U(\cx)$ and $\mathcal{F}_t$. Using a simple ordering of the $\bsx$ and appropriate change of notation we have,
\begin{align}
\label{eq:lambda_min_bound}
\lambda_{\min} \left(\frac{\boldsymbol{A}}{t}\right) & \geq \lambda_{\min} \left(\frac{1}{t} \sum_{i = 1}^{t\xi} \phi^t(\bsz^i) \phi^t(\bsz^i)^T + \frac{1}{t} \sum_{i = t\xi + 1}^{t} \phi^t(\bsx^i) \phi^t(\bsx^i)^T\right) \nonumber \\
&\geq \xi\lambda_{\min} \left(\frac{1}{t\xi} \sum_{i = 1}^{t\xi} \phi^t(\bsz^i) \phi^t(\bsz^i)^T\right) = \xi \lambda_{(m_t)} \left(K_{t\xi}^{m_t}\right)
\end{align}
where $\lambda_{(m_t)}$ denotes the $m_t$-th largest eigenvalue and $K_{t\xi}^{m_t}$ is the $t\xi \times t\xi$ matrix, whose $(i,j)$ entry is the $m_t$ level approximation of $k_{\eta_t}$. That is,
\begin{align*}
\left[K_{t\xi}^{m_t}\right]_{i,j} = \frac{1}{t\xi} k_{\eta_t}^{m_t} (\bsx_i, \bsx_j) =  \frac{1}{t\xi} \sum_{\ell=1}^{m_t} \lambda_{\ell}^t \psi_\ell^t( \bsx_i) \psi_\ell^t( \bsx_j).
\end{align*}
Now note that using the finite sample error bounds from \cite{braun2006accurate} we have,
\begin{align*}
\lambda_{(m_t)} \left(K_{t\xi}^{m_t}\right) \geq \lambda_{m_t}^t \left( 1 - C(t\xi, m_t)\right).
\end{align*}
Now from the results in \cite{braun2006accurate} if the kernel has bounded eigenfunctions, i.e., $|\psi_i(\bsx) | \leq M$, then for $\epsilon_1 > 0$, we have with probability larger than $1 - \epsilon_1$,
\begin{align*}
C(t\xi, m_t) < 
M^2 m_t \sqrt{ \frac{2}{t\xi} \log \frac{m_t(m_t+1)}{\epsilon_1}}.  
\end{align*}
On the other hand, if the kernel is bounded, i.e. $k(\bsx, \bsx) \leq M$ then with probability larger than $1 - \epsilon_1$
\begin{align*}
C(t\xi, m_t) < 
m_t \sqrt{ \frac{2 M }{t\xi \lambda_{m_t}^t} \log \frac{2m_t(m_t+1)}{\epsilon_1}}+ \frac{4M}{3 t\xi \lambda_{m_t}^t}  \log \frac{2m_t(m_t+1)}{\epsilon_1}.
\end{align*}
Choosing $\epsilon_1 = \epsilon^*/ t^{1+\delta}$ for some $\delta > 0$, we see that in both cases,
\begin{align*}
C(t\xi, m_t) = O\left(\sqrt{\frac{\log t}{t} }\right).
\end{align*}
Thus, if we denote an event as, 
\begin{align*}
E_t = \left\{\lambda_{(m_t)} \left(K_{t\xi}^{m_t}\right) \geq \lambda_{m_t}^t \left( 1 - c\sqrt{\frac{\log t}{t} }\right) \right\},
\end{align*}
then, $P(E_t^c) \leq \epsilon^*/t^{1+\delta}$ and hence, $\sum_{t=1}^\infty P(E_t^c) < \infty.$ Thus,  
by the Borel Cantelli Lemma, 
$P(E_t^c \text{ occurs infinitely often }) = 0$. Thus, outside a set of measure zero, for any $\omega \in \Omega$, and exists a $t(\omega)$ such that for all $t > t(\omega)$,
$$\lambda_{(m_t)} \left(K_{t\xi}^{m_t}\right) \geq \lambda_{m_t}^t \left( 1 - c\sqrt{\frac{\log t}{t} }\right).$$
Now for large enough $t$, $\lambda_{m_t}^t \rightarrow \lambda_{m^*} > 0$, where $\lambda_{m^*}$ is the $m^*$ largest eigenvalue of the optimal kernel $k_{\eta^*}$. If, $k_{\eta^*}$ has finitely many positive eigenvalues, then $m^*$ denotes the index of the smallest positive value. Otherwise $m^*$ is some finite large integer. Moreover, the multiplier term can be bounded by a constant $c^*$. Thus, outside a set of measure zero, for any $\omega \in \Omega$, there exists a $t(\omega)$ such that for all  $t > t(\omega)$ 
$$\lambda_{(m_t)} \left(K_{t\xi}^{m_t}\right) \geq c^* \lambda_{m^*}.$$ 
Hence, there exists a constant $c$ such that 
$$\liminf_{t \rightarrow \infty} \lambda_{\min} \left(\frac{\boldsymbol{A}}{t}\right) > \xi c > 0 \;\;\; \text{ almost surely.}$$

Similarly for the upper bound, we see for $t$ large enough,
\begin{align}
\label{eq:lambda_max_bound}
\lambda_{\max} \left(\frac{\boldsymbol{A}}{t}\right) &= \lambda_{\max}\left(\frac{\xi}{t\xi} \sum_{i = 1}^{t\xi} \phi^t(\bsz^i) \phi^t(\bsz^i)^T + \frac{1 - \xi}{t(1-\xi)} \sum_{i = t\xi + 1}^{t} \phi^t(\bsx^i) \phi^t(\bsx^i)^T + \frac{\sigma_t^2}{t}\boldsymbol{I}\right) \nonumber \\
& \leq \lambda_{\max}\left(\frac{\xi}{t\xi} \sum_{i = 1}^{t\xi} \phi^t(\bsz^i) \phi^t(\bsz^i)^T \right)\nonumber \\ &\qquad +   \lambda_{\max}\left(\frac{1 - \xi}{t(1-\xi)} \sum_{i = t\xi + 1}^{t} \phi^t(\bsx^i) \phi^t(\bsx^i)^T\right) + 1,
\end{align}
where we have used the fact that $\sigma_t^2/t \leq 1$ for large $t$. This is because for large $t$, $\sigma_t$ converges to $\sigma^*$ by the consistency of the maximum likelihood estimator. Now consider the second term.
\begin{align*}
 \lambda_{\max}\left(\frac{1 - \xi}{t(1-\xi)} \sum_{i = t\xi + 1}^{t} \phi^t(\bsx^i) \phi^t(\bsx^i)^T\right)  &= \frac{1 - \xi}{t(1-\xi)} \max_{\norm{x} = 1} \sum_{i = t\xi + 1}^t x^T\phi^t(\bsx^i) \phi^t(\bsx^i)^Tx \\
& \leq \frac{1 - \xi}{t(1-\xi)} \max_{\norm{x} = 1} \sum_{i = t\xi + 1}^t \norm{x}^2  \norm{\phi^t(\bsx^i)}^2  \leq (1 - \xi) \alpha,
\end{align*}
where we have used the Cauchy-Schwarz inequality and the regularity conditions on the  kernel which gives us,
$$\norm{\phi^t(\bsx^i)}^2 = k_{\eta_t}^{m_t}(\bsx^i, \bsx^i) \leq k_{\eta_t}(\bsx^i, \bsx^i) = \alpha \qquad \text{ for all } i.$$ 
Lastly, to control the first term we have 
\begin{align*}
\lambda_{\max}\left(\frac{1}{t\xi} \sum_{i = 1}^{t\xi} \phi^t(\bsz^i) \phi^t(\bsz^i)^T \right) = \lambda_{(1)} \left( K_{t\xi}^{mt} \right).
\end{align*}
Similar to the above proof and using the results from \cite{braun2006accurate}, we have with probability larger than $1 - \epsilon^*/t^{1+\delta}$,
\begin{align*}
\lambda_{(1)} \left(K_{t\xi}^{m_t}\right) \leq \lambda_{1}^t \left( 1 + c\sqrt{\frac{\log t}{t} } \right). 
\end{align*}
Applying the Borel-Cantelli Lemma, we have outside a set of measure zero, for any $\omega \in \Omega$, there exists a $t(\omega)$ such that for all  $t > t(\omega)$ 
\begin{align*}
\lambda_{(1)} \left(K_{t\xi}^{m_t}\right) \leq \lambda_{1} \tilde{c}^*, 
\end{align*}
where $\lambda_1$ denotes the maximum eigenvalue of the optimal kernel $k_{\eta^*}$.
Thus, there exists a constant $C$, such that
\begin{align*}
\limsup_{t \rightarrow \infty} \lambda_{\max} \left(\frac{\boldsymbol{A}}{t}\right) \leq \xi C + \alpha (1 - \xi) + 1 \;\;\; \text{ almost surely,}
\end{align*}
which completes the proof of the lemma.
\hfill$\blacksquare$

\subsection*{Proof of Lemma \ref{lemma:prod_eigen}}
Observe that,
\begin{align*}
\lambda_{\max} \left(\left(\frac{\boldsymbol{A}}{t}\right)^{-1} \frac{\boldsymbol {\Phi}^T  \boldsymbol {\Phi} }{t} \left(\frac{\boldsymbol{A}}{t}\right)^{-1}\right) 
&\leq  \lambda_{\max} \left(\frac{\boldsymbol{A}}{t}\right)^{-1} + \lambda_{\max} \left[\left(\frac{\boldsymbol{A}}{t}\right)^{-1} \boldsymbol{E}\right] 
\end{align*}
where $\boldsymbol{E} = \boldsymbol {\Phi}^T  \boldsymbol {\Phi} \boldsymbol{A}^{-1} - \boldsymbol{I}$. Now, it is easy to see that $\boldsymbol{E}$ is a negative definite matrix. Let $\xi_1^t \geq, \ldots, \geq \xi_{m_t}^t$ denote the eigenvalues of $ \boldsymbol {\Phi}^T  \boldsymbol {\Phi} /t$. Thus, using the spectral expansion there exists orthonormal eigenvectors $u_i$ such that,
\begin{align*}
\boldsymbol {\Phi}^T  \boldsymbol {\Phi} \boldsymbol{A}^{-1} =  \frac{\boldsymbol {\Phi}^T  \boldsymbol {\Phi}}{t} \left(\frac{\boldsymbol{A}}{t}\right)^{-1}=\sum_{i=1}^{m_t} \frac{\xi_i^t}{\xi_i^t + \sigma_t^2/t} u_iu_i^T \preccurlyeq \sum_{i=1}^{m_t} u_iu_i^T = \boldsymbol{I},
\end{align*}
where $\boldsymbol{A} \preccurlyeq \boldsymbol{B}$ implied, $\boldsymbol{B} - \boldsymbol{A}$ is positive definite. Thus, we get that $\boldsymbol{E}$ is negative definite. Therefore, using this and Lemma \ref{lemma:eigen} we have
\begin{align*}
\lambda_{\max} \left(\left(\frac{\boldsymbol{A}}{t}\right)^{-1} \frac{\boldsymbol {\Phi}^T  \boldsymbol {\Phi} }{t} \left(\frac{\boldsymbol{A}}{t}\right)^{-1}\right)  &\leq  \lambda_{\max} \left(\frac{\boldsymbol{A}}{t}\right)^{-1} - \lambda_{\min} \left[-\left(\frac{\boldsymbol{A}}{t}\right)^{-1} \boldsymbol{E}\right]  \\
&\leq \lambda_{\max} \left(\frac{\boldsymbol{A}}{t}\right)^{-1} \leq c,
\end{align*}
where we have used the result that for two positive definite matrices $\boldsymbol{A}, \boldsymbol{B},$ $\lambda_{\min}(\boldsymbol{AB}) \geq \lambda_{\min}(\boldsymbol{A})\lambda_{\min}(\boldsymbol{B}) > 0.$ This completes the proof of the Lemma.
\hfill$\blacksquare$

\subsection*{Proof of Lemma \ref{lemma:main_assumption}}

As in the proof of Lemma \ref{lemma:prod_eigen}, let $\xi_1^t \geq, \ldots, \geq \xi_{m_t}^t$ denote the eigenvalues of $ \boldsymbol {\Phi}^T  \boldsymbol {\Phi}/t $. Thus, using the spectral decomposition, 
\begin{align*}
\bignorm{ \left(\frac{\boldsymbol{A}}{t}\right)^{-1} \frac{\boldsymbol {\Phi}^T  \boldsymbol {\Phi}}{t} - \boldsymbol{I}} &= \sqrt{\lambda_{\max}\left[\left( \left(\frac{\boldsymbol{A}}{t}\right)^{-1} \frac{\boldsymbol {\Phi}^T  \boldsymbol {\Phi}}{t} - \boldsymbol{I} \right)^T \left( \left(\frac{\boldsymbol{A}}{t}\right)^{-1} \frac{\boldsymbol {\Phi}^T  \boldsymbol {\Phi}}{t} - \boldsymbol{I} \right) \right]}\\
&=  \sqrt{\lambda_{\max}\left[\left( \frac{\boldsymbol {\Phi}^T  \boldsymbol {\Phi}}{t} \left(\frac{\boldsymbol{A}}{t}\right)^{-1}  - \boldsymbol{I} \right) \left( \left(\frac{\boldsymbol{A}}{t}\right)^{-1} \frac{\boldsymbol {\Phi}^T  \boldsymbol {\Phi}}{t} - \boldsymbol{I} \right) \right]}\\
&= \sqrt{\lambda_{\max}\left( \frac{\boldsymbol {\Phi}^T  \boldsymbol {\Phi}}{t} \left(\frac{\boldsymbol{A}}{t}\right)^{-2} \frac{\boldsymbol {\Phi}^T  \boldsymbol {\Phi}}{t}  - \frac{\boldsymbol {\Phi}^T  \boldsymbol {\Phi}}{t} \left(\frac{\boldsymbol{A}}{t}\right)^{-1} - \left(\frac{\boldsymbol{A}}{t}\right)^{-1} \frac{\boldsymbol {\Phi}^T \boldsymbol {\Phi}}{t} +  \boldsymbol{I} \right)} \\
&= \sqrt{\lambda_{\max}\left[ \sum_{i=1}^{m_t} \left( \left(\frac{\xi_i^t}{\xi_i^t + \sigma_t^2/t}\right)^2 - \frac{2 \xi_i^t}{\xi_i^t + \sigma_t^2/t} + 1 \right) u_i u_i^T \right]} \\
&= \sqrt{\max_{i = 1, \ldots, m_t} \left(1 - \frac{\xi_i^t}{\xi_i^t + \sigma_t^2/t} \right)^2}.
\end{align*}
Now for each $i$, and large enough $t$, it easily follows from Lemma \ref{lemma:eigen} that the eigenvalues $\xi_i^t$ are bounded. Specifically,
$$0 < c \leq   \lambda_{\min}(\boldsymbol {\Phi}^T  \boldsymbol {\Phi}/t) \leq \xi_i^t \leq  \lambda_{\max}(\boldsymbol {\Phi}^T  \boldsymbol {\Phi}/t)  \leq C < \infty,$$
and $\sigma_t \rightarrow \sigma^*$ almost surely by the consistency of the maximum likelihood estimator. Thus, we get for some constant $c$ and large enough $t$,
\begin{align*}
\bignorm{ \left(\frac{\boldsymbol{A}}{t}\right)^{-1} \frac{\boldsymbol {\Phi}^T  \boldsymbol {\Phi}}{t} - \boldsymbol{I}} = \norm{\boldsymbol{A}^{-1} \boldsymbol {\Phi}^T  \boldsymbol {\Phi} - \boldsymbol{I}} \leq \frac{c}{t}
\end{align*}
which completes the first result. To prove the second result, note that using Lemma \ref{lemma:prod_eigen},\begin{align*}
\norm{\boldsymbol{A}^{-1} \boldsymbol {\Phi}^T}& = \sqrt{\lambda_{\max} ( \boldsymbol {\Phi} \boldsymbol{A}^{-1}\boldsymbol{A}^{-1} \boldsymbol {\Phi}^T)} = \sqrt{\lambda_{\max} (\boldsymbol{A}^{-1} \boldsymbol {\Phi}^T  \boldsymbol {\Phi} \boldsymbol{A}^{-1})} \\
&= \sqrt{\frac{1}{t}\lambda_{\max} \left(\left(\frac{\boldsymbol{A}}{t}\right)^{-1} \frac{\boldsymbol {\Phi}^T  \boldsymbol {\Phi} }{t} \left(\frac{\boldsymbol{A}}{t}\right)^{-1}\right)} \\
& \leq \frac{c}{\sqrt{t}},
\end{align*}
which completes the proof.
\hfill$\blacksquare$

\subsection*{Proof of Lemma \ref{lemma:numerator_bound}}
Using Assumption \ref{assu:func_norm} and Lemma \ref{lemma:main_assumption} observe that,
\begin{align*}
\norm{\bsu^T \boldsymbol{A}^{-1} \boldsymbol {\Phi}^T \bsf - \bsu^T \boldsymbol{A}^{-1} \boldsymbol {\Phi}^T\boldsymbol {\Phi} \btheta^* } &\leq \norm{\bsu}\norm{ \boldsymbol{A}^{-1} \boldsymbol {\Phi}^T } \norm{ \bsf - \boldsymbol {\Phi} \btheta^* }\\
& \leq 2\sqrt{\alpha t}  \delta_0(t) \norm{ \boldsymbol{A}^{-1} \boldsymbol {\Phi}^T } \\
&\leq c \delta_0(t),
\end{align*}
where we have used,
\begin{align}
\label{eq:norm_u}
\norm{\bsu}^2 = \norm{\phi^t(\bsx^*) - \phi^t(\bsx)}^2 \leq \left( \norm{\phi^t(\bsx^*)} + \norm{\phi^t(\bsx)}\right)^2 \leq 4\alpha,
\end{align}
and $\norm{\phi^t(\bsx)}^2 = k_{\eta_t}^{m_t}(\bsx, \bsx) \leq k_{\eta_t}(\bsx, \bsx) = \alpha.$ Thus we get, 
\begin{align}
\label{eq:mu2bound} 
\bsu^T \boldsymbol{A}^{-1} \boldsymbol {\Phi}^T \bsf &\geq \bsu^T \boldsymbol{A}^{-1} \boldsymbol {\Phi}^T \boldsymbol {\Phi} \btheta^* -  c\delta_0(t) \nonumber \\
&= \bsu^T  \btheta^* - \bsu^T \boldsymbol{E}\btheta^*  - c\delta_0(t).
\end{align}
We can now bound each of the terms on the right hand side of \eqref{eq:mu2bound}. Note that using Lemma \ref{lemma:main_assumption},
\begin{align}
\label{eq:bound_term2}
\norm{ \bsu^T \boldsymbol{E}\btheta^*} \leq \norm{\bsu} \norm{\boldsymbol{E}} \norm{\btheta^*} \leq 2\sqrt{\alpha} c M \frac{1}{\sqrt{t}}. 
\end{align}
Combining all the above bounds and using Assumption \ref{assu:func_norm} we have for large enough $t$,
\begin{align*}
\bsu^T \boldsymbol{A}^{-1} \boldsymbol {\Phi}^T \bsf &\geq \bsu^T  \btheta^* - c\left(\frac{1}{\sqrt{t}} + \delta_0(t)\right) \geq f(\bsx^*) - f(\bsx) - c\left(\frac{1}{\sqrt{t}} + \delta_0(t)\right)\\
&\geq \delta_\epsilon  - c\left(\frac{1}{\sqrt{t}} + \delta_0(t)\right) \geq \frac{\delta_\epsilon}{2},
\end{align*}
where the last inequality follows for large enough $t$ since $\lim_{t\rightarrow \infty}\delta_0(t) =0$ by Assumption \ref{assu:func_norm}. This completes the proof.
\hfill$\blacksquare$

\subsection*{Proof of Lemma \ref{lemma:chi_exp}}
Using Lemma 1 from \cite{laurent2000adaptive}, we have if $Z$ follows a chi-square distribution with $m$ degrees of freedom, then for any positive $x$,
\begin{align*}
P( Z \geq m + 2\sqrt{mx} + 2x) \leq \exp(-x).
\end{align*}
Now let $\delta = 2\sqrt{mx} + 2x$. Then, using the change of variables we have
$$x = -\frac{1}{2} \left(\delta + m + \sqrt{2\delta m + m^2}\right).$$
Plugging these in we get,
$$P( Z \geq m + \delta) \leq \exp\left(-\frac{1}{2} \left(\delta + m + \sqrt{2\delta m + m^2}\right)\right).$$
\hfill$\blacksquare$

\subsection*{Proof of Lemma \ref{lemma:prob_bound}}
For any $\epsilon' < \delta_\epsilon/4 $ and for any $\bsx$ such that $\norm{\bsx - \bsx^*} > \epsilon$, we have
\begin{align*}
P \big( \phi^t(\bsx^*)^T\btheta^t &< \phi^t(\bsx)^T\btheta^t + \epsilon' \big) \\
&= E_{\phi^t, D_{t}} \left( P \left( \phi^t(\bsx^*)^T\btheta^t < \phi^t(\bsx)^T\btheta^t + \epsilon' \bigg| \phi^t, D_{t}\right) \right).
\end{align*}
For notational simplicity we hide the variables we are conditioning on, specifically, $\phi^t, D_{t}$. Moreover, let $\bsv$ denote $\phi^t(\bsx) - \phi^t(\bsx^*)$. Under this simplified notation, let us define $X = \bsv^T\btheta^t$, which, given $\phi^t, D_{t}$, follows $N(\mu, \gamma^2)$ with $\mu =  \bsv^T \boldsymbol{A}^{-1} \boldsymbol {\Phi}^T  \boldsymbol{y}$ and $\gamma^2 = \sigma_t^2 \bsv^T  \boldsymbol{A}^{-1} \bsv$. Thus, we have
\begin{align}
\label{eq:first_split}
E \left( P \left( \bsv^T\btheta^t > - \epsilon' \right) \right) \leq E \left(\exp\left(-\frac{(\mu + \epsilon')^2}{2\gamma^2}\right)\right)  + P(\mu + \epsilon'> 0),
\end{align}
where the last inequality follows by conditioning on the sign on $\mu + \epsilon'$ and appropriately applying the tail bound for Gaussian random variables. Now, we separately consider the two terms in \eqref{eq:first_split}. For the first term, conditioning on $\bsx^0, \ldots, \bsx^{t-1}$ and $\phi^t$ define,
\begin{align*}
\zeta  &= \frac{\mu + \epsilon'}{\gamma} \bigg| \{\bsx^i\}_{i=0}^{t-1} , \phi^t \sim N(\tilde{\mu}, \tilde{\sigma}^2) \qquad \text { where }\\
\tilde{\mu} &=  \frac{\bsv^T \boldsymbol{A}^{-1} \boldsymbol {\Phi}^T \bsf + \epsilon'}{\sigma_t \sqrt{ \bsv^T  \boldsymbol{A}^{-1} \bsv}}\qquad \text { and } \qquad
\tilde{\sigma}^2 = \frac{\bsv^T \boldsymbol{A}^{-1} \boldsymbol {\Phi}^T\boldsymbol {\Phi} \boldsymbol{A}^{-1}\bsv}{\bsv^T  \boldsymbol{A}^{-1} \bsv},
\end{align*}
where $\bsf = (f(\bsx^0), \ldots, f(\bsx^{t-1}))^T$. Thus using the above notation we can write the first term as 
\begin{align*}
E \left(\exp\left(-\frac{(\mu+ \epsilon')^2}{2\gamma^2}\right)\right) &= E_{\bsx, \phi^t}\left( E\left(\exp\left(-\frac{\tilde{\sigma}^2}{2}\frac{\zeta^2}{\tilde{\sigma}^2}\right)\right)\right) \\
&= E_{\bsx, \phi^t}\left( \frac{1}{\sqrt{1 - 2\tau}}\exp\left( \frac{\lambda \tau}{1 - 2\tau} \right)  \right),
\end{align*}
where the last equality follows from the moment generating function of a non-central chi-squares distribution with non-centrality parameter $\lambda = \tilde{\mu}^2$ and $\tau = -\tilde{\sigma}^2/2$. Thus, we have,
\begin{align*}
E \left(\exp\left(-\frac{(\mu+\epsilon')^2}{2\gamma^2}\right)\right) &= E_{\bsx, \phi^t} \left( \frac{1}{\sqrt{1 + \tilde{\sigma}^2}} \exp\left( \frac{-\tilde{\sigma}^2 \tilde{\mu}^2}{2 ( 1+ \tilde{\sigma}^2)}\right)\right) \\
&\leq E_{\bsx, \phi^t} \left( \exp\left( \frac{-\tilde{\sigma}^2 \tilde{\mu}^2}{2 ( 1+ \tilde{\sigma}^2)}\right)\right).
\end{align*}
To give a lower bound to $\tilde{\sigma}^2$, observe that  
\begin{align*}
\tilde{\sigma}^2 &\geq \frac{\lambda_{\min}\left(  \boldsymbol{A}^{-1} \boldsymbol {\Phi}^T\boldsymbol {\Phi} \boldsymbol{A}^{-1} \right)}{\lambda_{\max}\left(  \boldsymbol{A}^{-1} \right)} = \lambda_{\min}\left(  \boldsymbol{A}^{-1} \boldsymbol {\Phi}^T\boldsymbol {\Phi} \boldsymbol{A}^{-1} \right) \lambda_{\min}\left(  \boldsymbol{A} \right) \\
&= \lambda_{\min}\left( \frac{ \boldsymbol{A}^{-1} }{t}\boldsymbol {\Phi}^T\boldsymbol {\Phi} \boldsymbol{A}^{-1} \right) \lambda_{\min}\left(  \frac{\boldsymbol{A}}{t} \right) = 
\lambda_{\min}\left( \frac{\boldsymbol{A}}{t}^{-1}  \boldsymbol{E} + \frac{\boldsymbol{A}}{t}^{-1}\right) \lambda_{\min}\left(  \frac{\boldsymbol{A}}{t} \right)\\
&\geq \left[\lambda_{\min}\left(\frac{\boldsymbol{A}}{t}^{-1}\right) + \lambda_{\min}\left(\frac{\boldsymbol{A}}{t}^{-1}  \boldsymbol{E}\right)\right]  \lambda_{\min}\left(  \frac{\boldsymbol{A}}{t} \right), 
\end{align*}
where $\boldsymbol{E} = \boldsymbol {\Phi}^T\boldsymbol {\Phi} \boldsymbol{A}^{-1}  - \boldsymbol{I}$. From the proof of Lemma \ref{lemma:main_assumption}, $\boldsymbol{E}$ is a negative definite matrix. Thus, we can write,
\begin{align*}
\tilde{\sigma}^2 &\geq \left[\lambda_{\min}\left(\frac{\boldsymbol{A}}{t}^{-1}\right) - \lambda_{\max}\left(-\frac{\boldsymbol{A}}{t}^{-1} \boldsymbol{E}\right)\right]  \lambda_{\min}\left(  \frac{\boldsymbol{A}}{t} \right). 
\end{align*}
Moreover, using the results in \cite{bhatia2013matrix},
\begin{align*}
\lambda_{\max}\left(-\frac{\boldsymbol{A}}{t}^{-1} \boldsymbol{E}\right) \leq \lambda_{\max} \left(\frac{\boldsymbol{A}}{t}^{-1}\right) \lambda_{\max}(-\boldsymbol{E}).
\end{align*}
Thus, we have,
\begin{align*}
\tilde{\sigma}^2 \geq \lambda_{\min}\left(\frac{\boldsymbol{A}}{t}^{-1}\right) \lambda_{\min}\left(  \frac{\boldsymbol{A}}{t} \right)  - \lambda_{\max}\left(-\boldsymbol{E}\right)
= \frac{\lambda_{\min}\left(  \frac{\boldsymbol{A}}{t} \right)}{\lambda_{\max}\left(  \frac{\boldsymbol{A}}{t} \right)}  - \lambda_{\max}\left(-\boldsymbol{E}\right).
\end{align*}
Now using Lemma \ref{lemma:eigen} and Lemma \ref{lemma:main_assumption}, there exists a $c^1, c^2$ and $c^3$ such that for large enough $t$, 
\begin{align*}
\lambda_{\max}(\boldsymbol{-E}) \leq \frac{c^1}{t}, \;\; \lambda_{\min}\left(  \frac{\boldsymbol{A}}{t} \right) \geq c^2\text{ and } \lambda_{\max}\left(  \frac{\boldsymbol{A}}{t} \right) \leq c^3.
\end{align*}
Thus, for large enough $t$ there exists a $c_4 > 0$ such that
$$\tilde{\sigma}^2 \geq c_4.$$
Denoting, $c = 1/(2+2/c_4)$ we have
\begin{align*}
E \left(\exp\left(-\frac{(\mu + \epsilon')^2}{2\gamma^2}\right)\right) &\leq E_{\bsx, \phi^t} \left( \exp\left( -c \tilde{\mu}^2\right)\right).
\end{align*}
To give a lower bound to $\tilde{\mu}^2$, we separately bound the numerator and the denominator. Using Lemma \ref{lemma:numerator_bound} we can give a lower bound to numerator of $\tilde{\mu}^2$. Specifically, for large enough $t$ we get
\begin{align*}
\left( \bsv^T \boldsymbol{A}^{-1} \boldsymbol {\Phi}^T \bsf + \epsilon'\right)^2 = \left|\bsu^T \boldsymbol{A}^{-1} \boldsymbol {\Phi}^T \bsf - \epsilon'\right|^2 \geq \frac{\delta_\epsilon^2}{16},
\end{align*}
where $\bsu = -\bsv = \phi^t(\bsx^*) - \phi^t(\bsx)$.
We now give an upper bound on the denominator of $\tilde{\mu}^2$. Note that using \eqref{eq:norm_u}
$$\norm{\bsv}^2 = \norm{\bsu}^2 \leq 4\alpha.$$
Hence we have,
\begin{align*}
\frac{\sigma_t^2}{t} \bsv^T  \left(\frac{\boldsymbol{A}}{t}\right)^{-1} \bsv \leq \frac{4 \alpha  \sigma_t^2}{t} \lambda_{\max}\left(\frac{\boldsymbol{A}}{t}\right)^{-1} \leq \frac{c}{t}.
\end{align*}
where the last inequality follows from Lemma \ref{lemma:eigen} and the consistency of $\sigma_t$ to $\sigma^*$, for large enough $t$. Combining the bounds for the numerator and the denominator we get,
\begin{align*}
E \left(\exp\left(-\frac{(\mu + \epsilon')^2}{2\gamma^2}\right)\right) &\leq  E_{\bsx, \phi^t} \left( \exp\left( -c \delta_\epsilon^2 t\right)\right)  \\
&=  \exp\left( -c\delta_\epsilon^2 t\right). 
\end{align*}
Using a very similar proof technique 
as above, we can show, $$P(\mu + \epsilon' > 0) \leq \exp\left(- c \delta_\epsilon^2 t \right). $$
Here we use $c$ as a generic constant. Thus, there exists a $c$ such that for large enough $t$, we have 
$$P \left( \phi^t(\bsx^*)^T\btheta^t < \phi^t(\bsx)^T\btheta^t  + \epsilon' \right ) \leq 2\exp(-c \delta_\epsilon^2 t). $$
This completes the proof of the lemma.
\hfill$\blacksquare$

\subsection*{Proof of Lemma \ref{lemma:norm_bound}}
Let $\mu_{\phi^t, D_{t}} =  \boldsymbol{A}^{-1} \boldsymbol {\Phi}^T \boldsymbol{y}$, $\Sigma_{\phi^t, D_{t}} =  \sigma_t^2 \boldsymbol{A}^{-1}$ and 
$$L_t =\sqrt{\sigma_t^2 \left( \frac{m_t}{t} + \delta\right)\left(\frac{1}{c_1} + \frac{1}{c_2} \right) +  c_3 t}.$$ Then,
\begin{align}
\label{eq:norm_theta_bound1}
P&\left( \norm{\btheta^t} >L_t\right) \leq P\left(\norm{\btheta^t}^2 >  L_t^2 \right) \nonumber\\
&= E_{D_{t}, \phi^t}\left( P\left(\norm{\btheta^t}^2 - \norm{\mu_{\phi^t, D_{t}}}^2 > L_t^2 -  \norm{\mu_{\phi^t, D_{t}}}^2 \Bigg| D_{t}, \phi^t  \right)\right) \nonumber\\
&\leq E_{D_{t}, \phi^t}\left( P\left(\norm{\btheta^t - \mu_{\phi^t, D_{t}}}^2 > L_t^2 -  \norm{\mu_{\phi^t, D_{t}}}^2 \Bigg| D_{t}, \phi^t  \right)\right) \nonumber\\
&\leq E_{D_{t}, \phi^t}\left( P\left(\norm{\btheta^t - \mu_{\phi^t, D_{t}}}_{\Sigma_{\phi^t,D_{t}}^{-1}}^2 > \lambda_{\min}(\Sigma_{\phi^t,D_{t}}^{-1})\left(L_t^2 -  \norm{\mu_{\phi^t, D_{t}}}^2\right) \Bigg| D_{t}, \phi^t  \right)\right).
\end{align}
Now, we condition on 
\begin{align}
\label{eq:condition_1}
\norm{\mu_{\phi^t, D_{t}}}^2 \leq  c_3 t + \left(\frac{m_t}{t} + \delta\right)\frac{\sigma_t^2}{c_2}. 
\end{align}
Using \eqref{eq:condition_1} we have,
\begin{align*}
\lambda_{\min}(\Sigma_{\phi^t,D_{t}}^{-1})\left(L_t^2 -  \norm{\mu_{\phi^t, D_{t}}}^2\right) &\geq \lambda_{\min}(\Sigma_{\phi^t,D_{t}}^{-1}) \frac{(m_t + \delta t)\sigma_t^2}{c_1 t}  \\
& = \frac{t}{\sigma_t^2}\lambda_{\min}\left(\frac{\boldsymbol{A}}{t}\right)\frac{(m_t + \delta t)\sigma_t^2}{c_1 t}\\
& \geq m_t + \delta t.
\end{align*}
Plugging this into \eqref{eq:norm_theta_bound1} and using Lemma \ref{lemma:chi_exp} we get,
\begin{align}
\label{eq:step1}
P&\left( \norm{\btheta^t} >L_t\right) \leq \exp\left(-\frac{\delta t}{2}\right) + P\left(\norm{\mu_{\phi^t, D_{t}}}^2 \geq  c_3 t + \left(\frac{m_t}{t} + \delta\right)\frac{\sigma_t^2}{c_2}\right).
\end{align}
Here we have specifically used the fact that $m_t$ can grow as $O(t)$. If it grows faster than that, then the first above term will have a different rate by using Lemma \ref{lemma:chi_exp}.  

Similar to the above analysis if we let $\mu_{\phi^t} =  \boldsymbol{A}^{-1} \boldsymbol {\Phi}^T \bsf$, $\Sigma_{\phi^t} =  \sigma_t^2 \boldsymbol{A}^{-1} \boldsymbol {\Phi}^T \boldsymbol{\Phi} \boldsymbol{A}^{-1}$ and 
$$
\tilde{L}_t^2 =  c_3 t + \left(\frac{m_t}{t} + \delta\right)\frac{\sigma_t^2}{c_2},
$$
then 
\begin{align}
\label{eq:norm_theta_bound3}
P&\left(\norm{\mu_{\phi^t, D_{t}}}^2 \geq \tilde{L}_t^2\right) \nonumber \\
&\leq E_{X_{t-1}, \phi^t}\left( P\left(\norm{\mu_{\phi^t, D_{t}} - \mu_{\phi^t}}_{\Sigma_{\phi^t}^{-1}}^2 > \lambda_{\min}\left(\Sigma_{\phi^t}^{-1}\right)\left(\tilde{L}_t^2 -  \norm{\mu_{\phi^t}}^2\right) \Bigg| X_{t-1}, \phi^t  \right)\right).
\end{align}
Now note that using Lemma \ref{lemma:prod_eigen} and \eqref{eq:constant_def2} we have for large enough $t$, 
\begin{align*}
\lambda_{\min}(\Sigma_{\phi^t}^{-1}) \left(\tilde{L}_t^2 -  \norm{\mu_{\phi^t}}^2\right) &= \frac{1}{\sigma_t^2 \lambda_{\max}\left( \boldsymbol{A}^{-1} (\boldsymbol {\Phi}^T \boldsymbol{\Phi}) \boldsymbol{A}^{-1} \right)} \left(\frac{m_t}{t} + \delta\right)\frac{\sigma_t^2}{c_2}\\
&\geq m_t + \delta t.
\end{align*}
This gives us, 
\begin{align}
\label{eq:step2}
P&\left(\norm{\mu_{\phi^t, D_{t}}}^2 \geq \tilde{L}_t^2\right)  \leq 
\exp\left(-\frac{\delta t}{2}\right).
\end{align}
Using the \eqref{eq:step1} and \eqref{eq:step2} for large enough $t$ we get,
\begin{align*}
P\left(\norm{\btheta^t} > \sqrt{\sigma_t^2 \left( \frac{m_t}{t} + \delta\right)\left(\frac{1}{c_1} + \frac{1}{c_2} \right) +  c_3 t}\right) &\leq 2\exp\left(-\frac{\delta t}{2}\right).
\end{align*}
which completes the proof of the Lemma. 
\hfill$\blacksquare$

\end{document}